\documentclass[preprint,12pt]{elsarticle}

\usepackage{graphicx}
\usepackage{amssymb}
\usepackage{float}
\usepackage{lineno,hyperref}
\usepackage{amsmath,amsfonts,amsthm,amssymb,graphicx,subfigure,mathtools,tikz,hyperref,bm,blindtext}
\usepackage{diagbox}
\usepackage{geometry}
\usepackage[acronym]{glossaries}
\usepackage{multirow}
\usepackage{algorithm}
\usepackage{algpseudocode}

\usepackage{lineno}

\makeglossaries
\newglossaryentry{PDEs}
{
name=PDEs,
description={partial differential equations}
}

\newglossaryentry{B-PINNs}
{
name=B-PINNs,
description={Bayesian physics-informed neural networks}
}

\newglossaryentry{PINNs}
{
name=PINNs,
description={physics-informed neural networks}
}

\newglossaryentry{BNNs}
{
name=BNNs,
description={Bayesian neural networks}
}

\newglossaryentry{KL}
{
name=KL,
description={Karhunen–Lo\`eve expansion}
}

\newglossaryentry{DNF}
{
name=DNF,
description={deep normalizing flow}
}

\newglossaryentry{HMC}
{
name=HMC,
description={Hamiltonian Monte Carlo}
}

\newglossaryentry{VI}
{
name=VI,
description={variational inference}
}

\newglossaryentry{GPR}
{
name=GPR,
description={Gaussian process regression}
}

\journal{Journal Name}

\begin{document}

\begin{frontmatter}


\title{B-PINNs: Bayesian Physics-Informed Neural Networks for Forward and Inverse PDE Problems with Noisy Data}



\author[brown]{Liu Yang\fnref{1}}
\author[brown]{Xuhui Meng\fnref{1}}
\author[brown,PNNL]{George Em Karniadakis \fnref{2}}

\fntext[1]{The first two authors contributed equally to this work.}
\fntext[2]{Corresponding author: george\_karniadakis@brown.edu (George Em Karniadakis).}
\address[brown]{Division of Applied Mathematics, Brown University, Providence, RI 02906, USA}
\address[PNNL]{Pacific Northwest National Laboratory, Richland, WA 99354, USA}

\begin{abstract}

 We propose a {\it Bayesian physics-informed neural network} (B-PINN) to solve both forward and inverse nonlinear problems described by partial differential equations (PDEs) and noisy data. In this Bayesian framework, the Bayesian neural network (BNN) combined with a PINN for PDEs serves as the prior while the Hamiltonian Monte Carlo (HMC) or the variational inference (VI) could serve as an estimator of the posterior. B-PINNs make use of both physical laws and scattered noisy measurements to provide predictions and quantify the \textit{aleatoric uncertainty} arising from the noisy data in the Bayesian framework. Compared with PINNs, in addition to uncertainty quantification, B-PINNs obtain more accurate predictions in scenarios with large noise due to their capability of avoiding overfitting. We conduct a systematic comparison between the two different approaches for the B-PINN posterior estimation (i.e., HMC or VI), along with dropout used for quantifying uncertainty in deep neural networks. Our experiments show that HMC is more suitable than VI for the B-PINNs posterior estimation, while dropout employed in PINNs can hardly provide accurate predictions with reasonable uncertainty. Finally, we replace the BNN in the prior with a truncated Karhunen-Lo\`eve (KL) expansion combined with HMC or a deep normalizing flow (DNF) model as posterior estimators. The KL is as accurate as BNN and much faster but this framework cannot be easily extended to high-dimensional problems unlike the BNN based framework.
 
\end{abstract}


\end{frontmatter}


\glsadd{B-PINNs}
\glsadd{PINNs}
\glsadd{PDEs}
\glsadd{BNNs}
\glsadd{HMC}
\glsadd{KL}
\glsadd{VI}
\glsadd{DNF}
\glsadd{GPR}

\printglossary[nonumberlist]

\section{Introduction}
\label{sec:intro}
The state-of-the-art in data-driven modeling has advanced significantly recently in applications across different fields \cite{lecun2015deep,rudy2017data,mangan2017model,brunton2019data,berg2019data}, due to the rapid development of machine learning and explosive growth of available data collected from different sensors (e.g., satellites, cameras, etc.). In general, purely data-driven methods require a large amount of data in order to get accurate results \cite{raissi2019physics}. As a powerful alternative, recently the data-driven solvers for partial differential equations (PDEs) have drawn an increasing attention due to their capability to encode the underlying physical laws in the form of PDEs and give relatively accurate predictions for the unknown terms with limited data. In the first case we need ``big data" while in the second case we can learn from ``small data" as we explicitly utilize the physical laws or more broadly a parametrization of the physics.

Two typical approaches are the Gaussian processes regression (GPR) for PDEs \citep{raissi2017machine}, and the physics-informed neural networks (PINNs) \cite{raissi2019physics,lu2019deepxde}. Built upon the Bayesian framework with built-in mechanism for uncertainty quantification, GPR is one of the most popular data-driven methods. However, vanilla GPR has difficulties in handling the nonlinearities when applied to solve PDEs, leading to restricted applications. On the other hand, PINNs have shown effectiveness in both forward and inverse problems for a wide range of PDEs \cite{zhang2019quantifying,yang2020physics,meng2020composite,meng2019ppinn,mao2020physics}. However, PINNs are not equipped with built-in uncertainty quantification, which may restrict their applications, especially for  scenarios where the data are noisy.

\begin{figure}[h]
    \centering
    \includegraphics[width=0.7\textwidth]{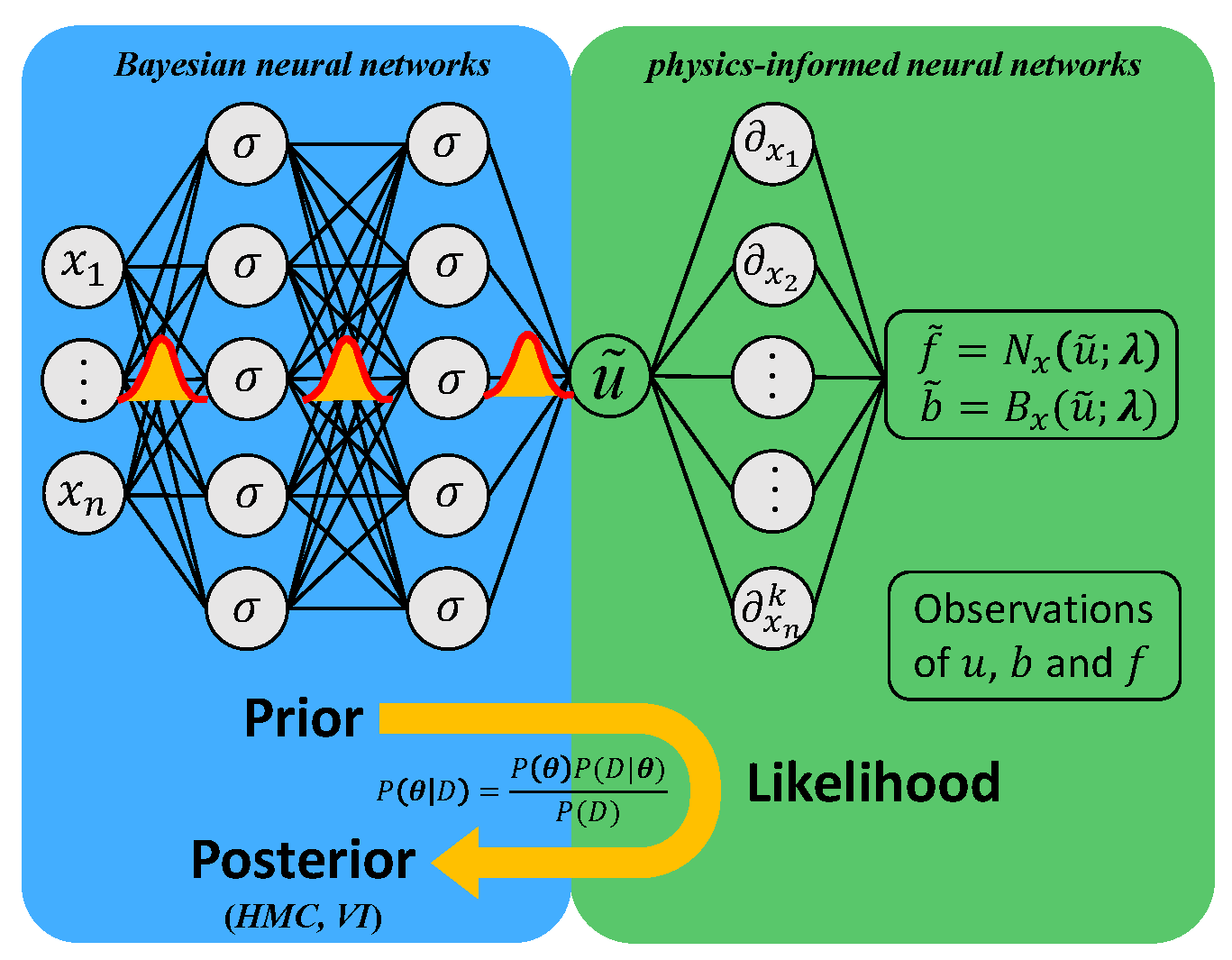}
    \caption{
    Schematic for the Bayesian physics-informed neural network (B-PINN). $P(\bm{\theta})$ is the prior for hyperparameters as well as the unknown terms in  PDEs, $P\left(\mathcal{D}|\bm{\theta} \right)$ represents the likelihood of observations (e.g., $u,~ b, ~f$), and $P(\bm{\theta}|\mathcal{D})$ is the posterior. The blue panel 
    represents the Bayesian neural network while the green panel represents the physics-informed part.
    }
    \label{fig:B-PINNs}
\end{figure}

In previous work, we use physics-informed generative adversarial networks to quantify parametric uncertainty \cite{yang2020physics} and also polynomial chaos expansions  in conjunction with dropout to quantify total uncertainty \cite{zhang2019quantifying}. In the present work, we propose a Bayesian physics-informed neural networks (B-PINN) to solve linear or nonlinear PDEs with noisy data, see Fig. \ref{fig:B-PINNs}. The uncertainties arising from the scattered noisy data could be naturally quantified due to the Bayesian framework \cite{luo2020bayesian}. B-PINNs consist of two parts: a parameterized surrogate model, i.e., a Bayesian neural network (BNN) with prior for the unknown terms in a PDE,  and an approach for estimating the posterior distributions of the parameters in the surrogate model. In particular, we employ the Hamiltonian Monte Carlo (HMC) \cite{neal2011mcmc,neal2012bayesian} or the variational inference (VI) \cite{graves2011practical,blundell2015weight} for estimation of the posterior distributions. In addition, we note that a non-Bayesian framework model, i.e., the dropout, has been used to quantify the uncertainty in deep neural networks, including the PINNs for solving PDEs \cite{zhang2019quantifying,gal2016dropout}. We will validate the proposed B-PINN method and conduct a systematic comparison with the dropout for both the forward and inverse PDE problems given noisy data.

In addition to BNNs, the Karhunen-Lo\`eve expansion is also a widely used representation of a stochastic process. As an illustration, we further test the case using the truncated Karhunen-Lo\`eve as the surrogate model while we use HMC or the deep normalizing flow (DNF) models \cite{yang2019potential} for estimating the posterior in the Bayesian framework.

The rest of the paper is organized as follows: In Sec. \ref{sec:bayesian}, we present the B-PINN algorithm for solving forward/inverse PDE problems with noisy data, including the BNNs for PDEs and posterior estimation methods, i.e., the HMC and VI, used in this paper. In Sec. \ref{sec:results}, we compare the performance of the B-PINNs and dropout on the tasks of function approximation, forward PDE problems, and inverse PDE problems.  In addition, we present comparisons between B-PINNs and PINNs as well as the KL for nonlinear forward/inverse PDEs in Secs. \ref{sec:pinn}-\ref{sec:kl}. We make a summary in Sec. \ref{sec:summary}. Furthermore, in \ref{sec:prior} we present a study on the priors of BNNs, and in \ref{sec:DNF} we give more details on the DNF models.

\section{B-PINNs: Bayesian Physics-informed Neural Networks}
\label{sec:bayesian}
We consider a general partial differential equation (PDE)  of the form 
\begin{equation}
\begin{aligned}
    \mathcal{N}_{\boldsymbol{x}}(u;\boldsymbol{\lambda}) &= f, \quad \boldsymbol{x}\in D, \\
    \mathcal{B}_{\boldsymbol{x}}(u;\boldsymbol{\lambda}) & = b, \quad \boldsymbol{x} \in \Gamma,
\end{aligned}
\end{equation}
where $\mathcal{N}_{\boldsymbol{x}}$ is a general differential operator, $D$ is the $d$-dimensional physical domain, $u = u(\bm{x})$ is the solution of the PDE, and $\boldsymbol{\lambda}$ is the vector of parameters in the PDE. 
Also, $f = f(\bm{x})$ is the forcing term, and $\mathcal{B}_{\boldsymbol{x}}$ is the boundary condition operator acting on the domain boundary $\Gamma$. In forward problems $\boldsymbol{\lambda}$ is prescribed, and hence our goal is to infer the distribution of $u$ at any $\boldsymbol{x}$. In inverse problems, $\boldsymbol{\lambda}$ is also to be inferred from the data.

We consider the scenario where our available dataset $\mathcal{D}$ are scattered noisy measurements of $u$, $f$ and $b$ from sensors:
\begin{equation}
    \begin{aligned}
    \mathcal{D} & = \mathcal{D}_u\cup\mathcal{D}_f\cup\mathcal{D}_b,
        \end{aligned}
\end{equation}
where$\mathcal{D}_u = \{(\boldsymbol{x}_{u}^{(i)},~ \bar{u}^{(i)})\}_{i=1}^{N_u}, ~\mathcal{D}_f  = \{(\boldsymbol{x}_{f}^{(i)}, ~\bar{f}^{(i)})\}_{i=1}^{N_f},
    ~\mathcal{D}_b = \{(\boldsymbol{x}_{b}^{(i)}, ~\bar{b}^{(i)})\}_{i=1}^{N_b}.$
We assume that the measurements are independently Gaussian distributed centered at the hidden real value, i.e.,
\begin{equation}
    \begin{aligned}
    \bar{u}^{(i)} &= u(\boldsymbol{x}_{u}^{(i)}) + \epsilon_u^{(i)}, \quad i = 1,2...N_u, \\
    \bar{f}^{(i)} &= f(\boldsymbol{x}_{f}^{(i)}) + \epsilon_f^{(i)}, \quad i = 1,2...N_f, \\
    \bar{b}^{(i)} &= b(\boldsymbol{x}_{b}^{(i)}) + \epsilon_b^{(i)}, \quad i = 1,2...N_b,
    \end{aligned}
\end{equation}
where $\epsilon_u^{(i)}$, $\epsilon_f^{(i)}$ and $\epsilon_b^{(i)}$ are independent Gaussian noises with zero mean. We also assume that the fidelity of each sensor is known, i.e., the standard deviations of $\epsilon_u^{(i)}$, $\epsilon_f^{(i)}$ and $\epsilon_b^{(i)}$ are known to be $\sigma_u^{(i)}$, $\sigma_f^{(i)}$ and $\sigma_b^{(i)}$, respectively. Note that the size of the noise could be different among measurements of different terms, and even between measurements of the same terms in the PDE.

We firstly consider the {\em forward} problem setup. The Bayesian framework starts from representing $u$ with a surrogate model $\tilde{u}(\boldsymbol{x}; \boldsymbol{\theta})$, where $\boldsymbol{\theta}$ is the vector of parameters in the surrogate model with a prior distribution $P(\boldsymbol{\theta})$. Consequently, $f$ and $b$ are represented by: 
\begin{equation}
\label{eqn:fb}
\begin{aligned}
    \tilde{f}(\boldsymbol{x}; \boldsymbol{\theta}) = \mathcal{N}_{\boldsymbol{x}}(\tilde{u}(\boldsymbol{x}; \boldsymbol{\theta});\boldsymbol{\lambda}), ~\tilde{b}(\boldsymbol{x}; \boldsymbol{\theta}) = \mathcal{B}_{\boldsymbol{x}}(\tilde{u}(\boldsymbol{x}; \boldsymbol{\theta}) ;\boldsymbol{\lambda}).
\end{aligned}
\end{equation}
Then, the likelihood can be calculated as:
\begin{equation}
\label{eqn:likelihood}
\begin{aligned}
    P(\mathcal{D}|\boldsymbol{\theta}) &= P(\mathcal{D}_u|\boldsymbol{\theta}) P(\mathcal{D}_f|\boldsymbol{\theta}) P(\mathcal{D}_b|\boldsymbol{\theta}), \\
     P(\mathcal{D}_u|\boldsymbol{\theta}) &= \prod_{i=1}^{N_u} \frac{1}{\sqrt{2\pi{\sigma_u^{(i)}}^2}}\exp \left(-\frac{(\tilde{u}(\boldsymbol{x}_{u}^{(i)}; \boldsymbol{\theta}) - \bar{u}^{(i)})^2}{2{\sigma_u^{(i)}}^2}\right), \\
     P(\mathcal{D}_f|\boldsymbol{\theta}) &= \prod_{i=1}^{N_f} \frac{1}{\sqrt{2\pi{\sigma_f^{(i)}}^2}}\exp \left(-\frac{(\tilde{f}(\boldsymbol{x}_{f}^{(i)}; \boldsymbol{\theta}) - \bar{f}^{(i)})^2}{2{\sigma_f^{(i)}}^2}\right), \\
    P(\mathcal{D}_b|\boldsymbol{\theta}) &= \prod_{i=1}^{N_b} \frac{1}{\sqrt{2\pi{\sigma_b^{(i)}}^2}}\exp \left(-\frac{(\tilde{b}(\boldsymbol{x}_{b}^{(i)}; \boldsymbol{\theta}) - \bar{b}^{(i)})^2}{2{\sigma_b^{(i)}}^2}\right). \\
\end{aligned}
\end{equation}
Finally, the posterior is obtained from Bayes' theorem:
\begin{equation}
\label{eqn:forwardpost}
\begin{aligned}
P(\boldsymbol{\theta}| \mathcal{D}) = \frac{ P(\mathcal{D}|\boldsymbol{\theta})P(\boldsymbol{\theta})}{P(\mathcal{D})} \simeq P(\mathcal{D}|\boldsymbol{\theta})P(\boldsymbol{\theta}),
\end{aligned}
\end{equation}
where ``$\simeq$'' represents equality up to a constant. Usually the calculation of $P(\mathcal{D})$ is analytically intractable, thus in practice we only have an unnormalized expression of $P(\boldsymbol{\theta}|\mathcal{D})$. To give a posterior $u$ at any $\boldsymbol{x}$, we can sample from $P(\boldsymbol{\theta}| \mathcal{D})$, denoted as $\{{\boldsymbol{\theta}}^{(i)}\}_{i=1}^M$, and then obtain statistics from samples $\{\tilde{u}(\boldsymbol{x}; {\boldsymbol{\theta}}^{(i)} ) \}_{i=1}^M$. We focus mostly on the mean and standard deviation of $\{\tilde{u}(\boldsymbol{x}; {\boldsymbol{\theta}}^{(i)} ) \}_{i=1}^M$, since the former represents the prediction of  $u(\boldsymbol{x})$ while the latter quantifies the uncertainty.

In the case of inverse problems, we can build the surrogate model for $u$ in the same way as above. However, apart from $\boldsymbol{\theta}$, we also need to assign a prior distribution for $\boldsymbol{\lambda}$, which could be independent of $P(\boldsymbol{\theta})$. The likelihood is the same as in Eq.~\ref{eqn:likelihood}, except that $P(\mathcal{D}|\boldsymbol{\theta})$, $P(\mathcal{D}_u|\boldsymbol{\theta})$, $P(\mathcal{D}_f|\boldsymbol{\theta})$ and $P(\mathcal{D}_b|\boldsymbol{\theta})$ should be replaced by 
$P(\mathcal{D}|\boldsymbol{\theta},\boldsymbol{\lambda})$, $P(\mathcal{D}_u|\boldsymbol{\theta},\boldsymbol{\lambda})$, $P(\mathcal{D}_f|\boldsymbol{\theta},\boldsymbol{\lambda})$ and $P(\mathcal{D}_b|\boldsymbol{\theta},\boldsymbol{\lambda})$, respectively.  Consequently, we should calculate the joint posterior of $[\boldsymbol{\theta}, \boldsymbol{\lambda}]$ as
\begin{equation}
\label{eqn:backwardpost}
\begin{aligned}
P(\boldsymbol{\theta}, \boldsymbol{\lambda} | \mathcal{D}) = \frac{ P(\mathcal{D}|\boldsymbol{\theta},\boldsymbol{\lambda})P(\boldsymbol{\theta},\boldsymbol{\lambda})}{P(\mathcal{D})} \simeq P(\mathcal{D}|\boldsymbol{\theta},\boldsymbol{\lambda})P(\boldsymbol{\theta},\boldsymbol{\lambda}) = P(\mathcal{D}|\boldsymbol{\theta},\boldsymbol{\lambda})P(\boldsymbol{\theta})P(\boldsymbol{\lambda}),
\end{aligned}
\end{equation}
where the last equality comes from the fact that the priors for $\boldsymbol{\theta}$ and $\boldsymbol{\lambda}$ are independent.

The parameter $\boldsymbol{\lambda}$ in the PDE is a vector in the above problem setup, however, we remark that the same framework could be applied in the cases where the parameter is a field or fields depending on $\boldsymbol{x}$, by representing the parameter vector with another surrogate model.

Since the forward problems and inverse problems are formulated in the same framework, in the following we will use $\boldsymbol{\theta}$ to represent the vector of all the unknown parameters in the surrogate models for the solutions and parameters. We denote the dimension of $\boldsymbol{\theta}$, i.e., the number of unknown parameters, as $d_{\boldsymbol{\theta}}$.

\subsection{Prior for Bayesian Physics-informed Neural Networks}
We consider a fully-connected neural network with $L \ge 1$ hidden layers as the surrogate model, see Fig. \ref{fig:B-PINNs}. Let us denote the input of the neural network as $\bm{x} \in R^{N_x}$, the output of the neural network as $\tilde{u} \in R$, and the $l$-th hidden layer as $\bm{z}_{l} \in R^{N_l}$ for $l = 1,2...N$. Then
\begin{equation}
    \begin{aligned}
    \bm{z}_{l} &= \phi(\bm{w}_{l-1}\bm{z}_{l-1} + \bm{b}_{l-1}), \quad l = 1, 2...L,\\
    \tilde{u} &= \bm{w}_{L}\bm{z}_{L} + \bm{b}_{L}, 
    \end{aligned}
\end{equation}
where $\bm{w}_l \in R^{N_{l+1}\times N_l}$ are the weight matrices, $b_l \in R^{N_{l+1}}$ are the bias vectors, $\phi$ is the nonlinear activation function, which is the hyperbolic tangent function in the present study, and $\bm{z}_0 = \bm{x}, ~N_0 = N_x$, and $N_{L+1} = 1$ for the convenience of notation. When using a neural network as a surrogate model, the unknown parameters $\bm{\theta}$ are the concatenation of all the weight matrices and bias vectors.

\begin{figure}
    \centering
    \subfigure[]{\label{fig:density_u}
    \includegraphics[width=0.31\textwidth]{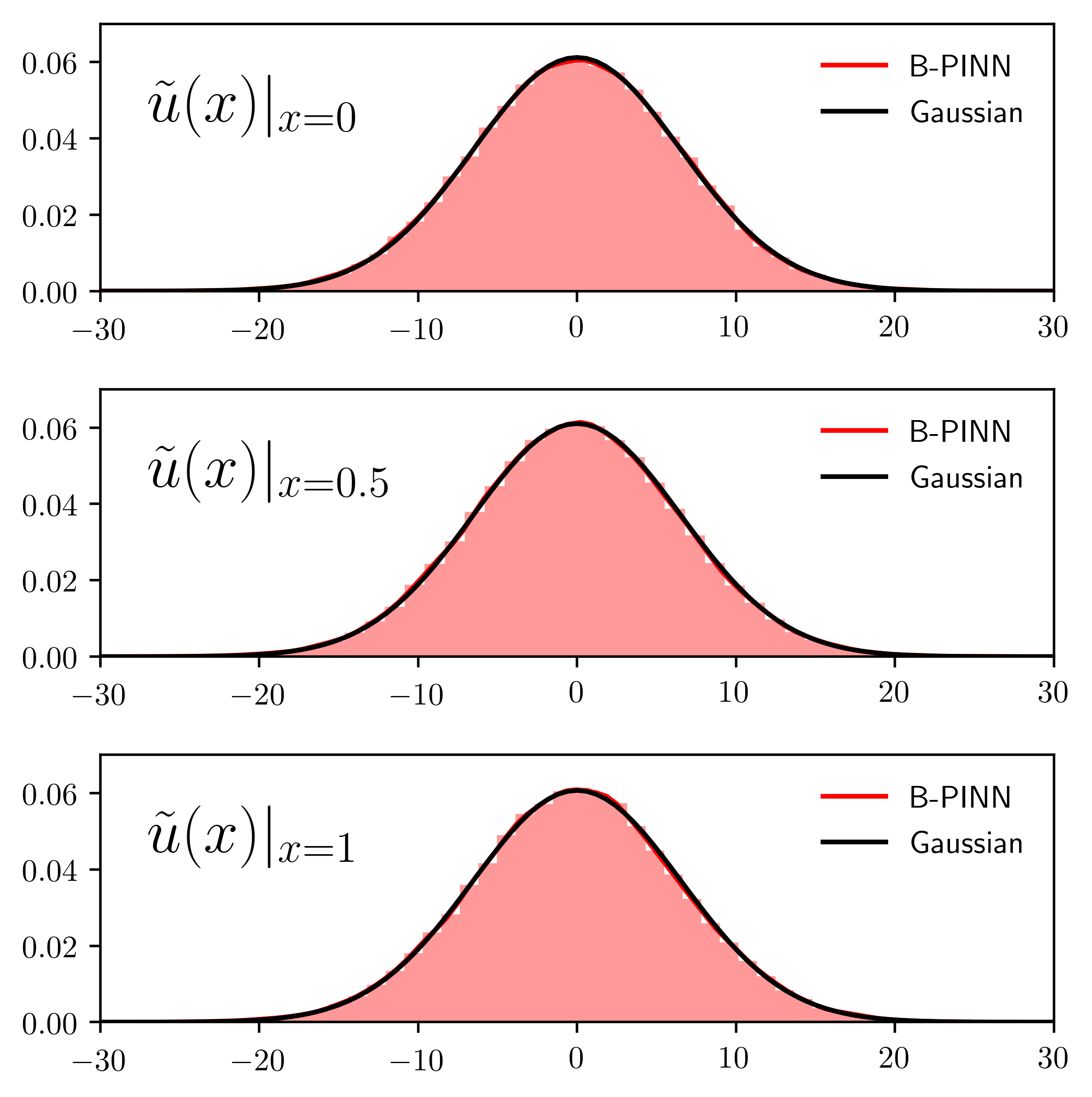}}
    \subfigure[]{\label{fig:density_f}
    \includegraphics[width=0.31\textwidth]{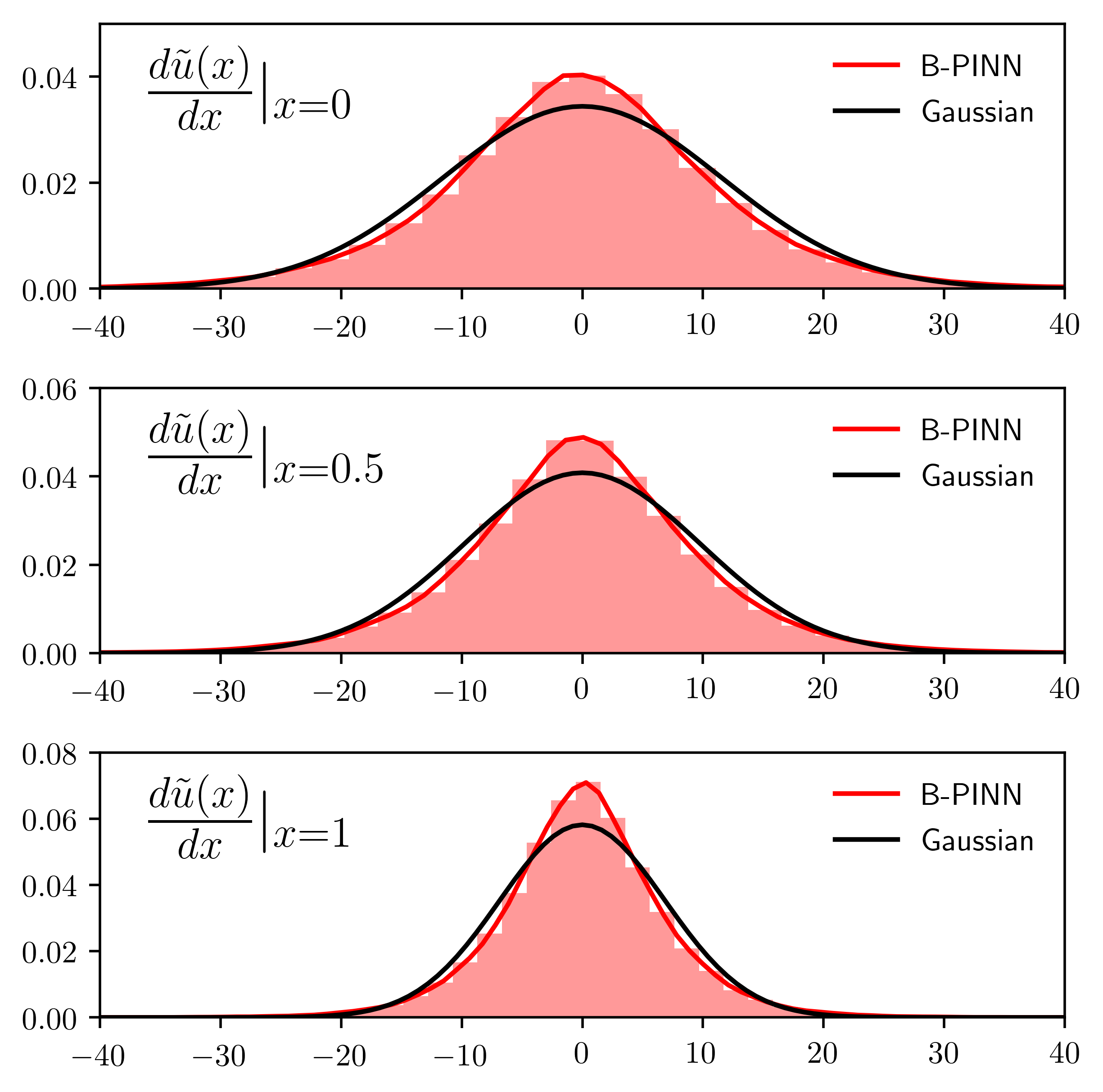}}
    \subfigure[]{\label{fig:density_ff}
    \includegraphics[width=0.31\textwidth]{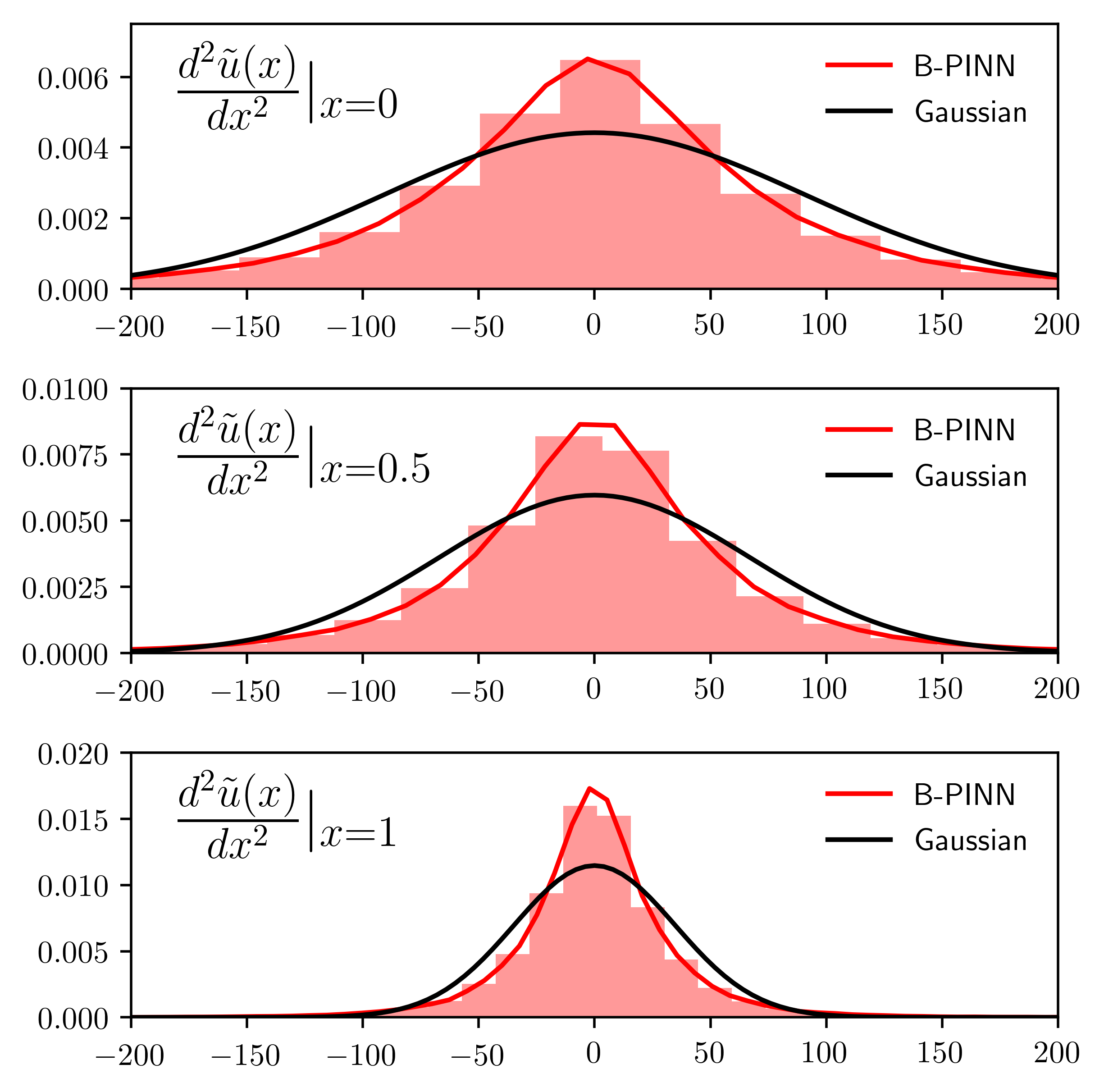}}
    \caption{Comparison between the B-PINNs' prior distributions and Gaussian distributions. The red lines and histograms represent the density of B-PINNs' outputs (a) $\tilde{u}(x)$, (b) $d\tilde{u}(x)/dx$, and (c) $d^2\tilde{u}(x)/dx^2$ at $x = 0$, $0.5$, and $1$. The black lines are the density functions of the corresponding Gaussian distributions with zero mean and the same standard deviations as the B-PINNs' outputs.}
    \label{fig:prior_density}
\end{figure}

In the application of Bayesian neural networks, a commonly used prior for $\bm{\theta}$ is that each component of $\bm{\theta}$ is an independent Gaussian distribution with zero mean, and the entries of $\bm{w}_l$ and $\bm{b}_l$ have the variances $\sigma_{w,l}$ and $\sigma_{b,l}$, respectively, for $l = 0,1...L$ \cite{neal2011mcmc,neal2012bayesian}. In this case, it can be shown that the prior of the function $\tilde{u}(\bm{x})$ is actually a Gaussian process as the width of hidden layers goes to infinity with $\sqrt{N_l}\sigma_{w,l}$ fixed for $l = 1,2...L$ \cite{neal2012bayesian,lee2017deep,pang2019neural}. 

However, we remark that due to the finite width of the neural networks, the derivatives of $\tilde{u}$ could be far from Gaussian processes, in contrast to the derivatives of a Gaussian process (under certain regularity constraints) which are also Gaussian processes. For example, in Fig.~\ref{fig:prior_density} we compare the prior of B-PINNs with Gaussian distributions, where $N_x = 1$, $L=2$, $N_1 = N_2 = 50, ~\sigma_{b,l} = \sigma_{w,l} = 1$, for $l = 0,1,2$. We can see that although the priors of $\tilde{u}(x)$ at various $x$ match the corresponding Gaussian distributions, the priors of $d\tilde{u}(x)/dx$ and $d^2\tilde{u}(x)/dx^2$ are not close to the Gaussian distributions.

\subsection{Posterior sampling approaches for Bayesian Physics-informed Neural Networks}
In this subsection, we introduce two approaches to sample from the posterior distribution of the parameters in B-PINNs: the Hamiltonian Monte Carlo (HMC) method and the variational inference (VI) method.
\subsubsection{Hamiltonian Monte Carlo (HMC) method}
Hamiltonian Monte Carlo, which is known as a golden approach for sampling from posterior distributions, is an efficient Markov Chain Monte Carlo (MCMC) method based on the Hamiltonian dynamics \cite{neal2011mcmc,neal2012bayesian,betancourt2017conceptual}. In this approach, we first simulate the Hamiltonian dynamics using numerical integration, which is then corrected by an Metropolis-Hastings acceptance step.

Suppose the target posterior distribution for $\bm{\theta}$ given a certain number of observations $\mathcal{D}$ is defined as 
\begin{align}
    P(\bm{\theta} | \mathcal{D}) \simeq \exp(-U(\bm{\theta})),
\end{align}
where
\begin{align}
    U(\theta) = - \ln P(\mathcal{D}|\bm{\theta}) - \ln P(\bm{\theta}).
\end{align}

To sample from the posterior, HMC first introduces an auxiliary momentum variable $\bm{r}$ to construct a Hamiltonian system 
\begin{align}
    H(\bm{\theta}, \bm{r}) = U(\bm{\theta}) + \frac{1}{2} \bm{r}^T \bm{M}^{-1} \bm{r},
\end{align}
 where $\bm{M}$ is a mass matrix, which is often set to be identity matrix, $\bm{I}$. Then HMC generates samples from a joint distribution of $(\bm{\theta}, \bm{r})$ as follows
 \begin{align}
     \pi(\bm{\theta}, \bm{r}) \thicksim \exp(-U(\bm{\theta}) - \frac{1}{2}\bm{r}^T \bm{M}^{-1} \bm{r}).
 \end{align}
As we simply discard the $\bm{r}$ samples, the $\bm{\theta}$ samples have marginal distribution $P(\bm{\theta} | \mathcal{D})$. 

Specifically, the samples are generated from the following Hamiltonian dynamics
\begin{subequations}\label{eq:hmc}
\begin{align}
    d \bm{\theta} &= \bm{M}^{-1} \bm{r} dt,\\
    d{\bm{r}} &= - \nabla U(\bm{\theta}) dt.
\end{align}
\end{subequations}
We use the leapfog method to discretize Eq. \eqref{eq:hmc}. Furthermore, to reduce the discretization error, 
we employ a Metropolis-Hastings step.  The details for implementing the HMC method are displayed in Algorithm  \ref{alg:hmc}.

\begin{algorithm}[H]
\caption{Hamiltonian Monte Carlo}
\label{alg:hmc}
\begin{algorithmic}
\Require initial states for $\bm{\theta}^{t_0}$ and time step size $\delta t$.
\For{$k=1,2...N$}
    \State Sample $\bm{r}^{t_{k-1}}$ from $\mathcal{N}(0, \bm{M})$,\;
    \State $(\bm{\theta}_0$, $\bm{r}_0) \leftarrow (\bm{\theta}^{t_{k-1}}, \bm{r}^{t_{k-1}})$.\;
 \For{$i=0,1...(L-1)$} 
    \State $\bm{r}_{i} \gets \bm{r}_{i} - \frac{\delta t}{2} \nabla U(\bm{\theta}_i)$,\;
    \State$\bm{\theta}_{i+1} \gets \bm{\theta}_{i} + \delta t \bm{M}^{-1} \bm{r}_{i}$,\;
    \State$\bm{r}_{i+1} \gets \bm{r}_{i} - \frac{\delta t}{2} \nabla U(\bm{\theta}_{i+1})$,\;
  \EndFor
\State Metropolis-Hastings step:\;
\State Sample $p$ from $\mbox{Uniform}[0, 1]$, \;
\State $\alpha \gets \mbox{min}\{ 1, \exp(H(\bm{\theta}_L, \bm{r}_L) - H(\bm{\theta}^{t_{k-1}}, \bm{r}^{t_{k-1}})) \}$.\;
  \If{$p \ge \alpha$}
  \State $\bm{\theta}^{t_{k}} \gets \bm{\theta}_L$,\;
  \Else
  \State $\bm{\theta}^{t_{k}} \gets \bm{\theta}^{t_{k-1}}$.\;
  \EndIf
\EndFor
\State Calculate $\{\tilde{u}(\boldsymbol{x},\boldsymbol{\theta}^{t_{N+1-j}})\}_{j=1}^M$ as samples of $u(\boldsymbol{x})$, similarly for other terms. \;
\end{algorithmic}
\end{algorithm}

\subsubsection{Variational Inference (VI) method}

In the variational learning, the posterior density of the unknown parameter vector $\bm{\theta} = (\theta_1, \theta_2...\theta_{d_{\bm{\theta}}})$, i.e., $P(\bm{\theta} | \mathcal{D})$, is approximated by another density function $Q(\bm{\theta}; \bm{\zeta})$ parameterized by $\bm{\zeta}$, which is restricted to a smaller family of distributions \cite{blundell2015weight,yao2019quality}. A commonly used form of $Q$ is a factorizable Gaussian distribution as follows: 
\begin{equation}
\label{eqn:VI_Q}
    Q(\bm{\theta}; \bm{\zeta}) = \prod_{i=1}^{d_{\bm{\theta}}} q(\theta_i; \zeta_{\mu,i}, \zeta_{\rho,i}),
\end{equation}
where $\bm{\zeta} = (\bm{\zeta}_{\mu},\bm{\zeta}_{\rho})$, $\bm{\zeta}_{\mu} = (\zeta_{\mu,1},\zeta_{\mu,2}...\zeta_{\mu,d_{\bm{\theta}}})$, $\bm{\zeta}_{\rho} = (\zeta_{\rho,1},\zeta_{\rho,2}...\zeta_{\rho,d_{\bm{\theta}}})$, and $q(\theta_i; \zeta_{\mu,i}, \zeta_{\rho,i})$ is the density of the one-dimensional Gaussian distribution with mean $\zeta_{\mu,i}$ and standard deviation $\ln(1+\exp(\zeta_{\rho,i}))$.

Different versions of VI have been developed \cite{graves2011practical,blundell2015weight}, and here we employ the relatively popular one developed in \cite{blundell2015weight}, which is also easy to implement. In this approach, we can tune $\bm{\zeta}$ to minimize
\begin{equation}
    \begin{aligned}
    D_{KL}(Q(\bm{\theta};\bm{\zeta})||P(\bm{\theta}|\mathcal{D})) \simeq \mathbb{E}_{\bm{\theta}\sim Q} [\ln Q(\bm{\bm{\theta}}; \bm{\zeta}) - \ln P(\bm{\theta}) - \ln P(\mathcal{D} | \bm{\theta})],
    \end{aligned}
\end{equation} 
where $D_{KL}$ denotes the Kullback-Leibler divergence.

Here we employ the Adam optimizer \cite{kingma2014adam} to train $\bm{\zeta}$. The detailed algorithm is given in Algorithm \ref{alg:vi}. We refer the readers to \cite{blundell2015weight} for more details on the  variational inference.

\begin{algorithm}[H]
\caption{Variational inference}
\label{alg:vi}
\begin{algorithmic}
\Require an initial state for $\bm{\zeta}$.
\For{$k=1,2...N$}
\State Sample $\{\bm{z}^{(j)}\}_{j=1}^{N_z}$ independently from $\mathcal{N}(\bm{0}, \bm{I}_{d_{\bm{\theta}}})$.\;
\State $\bm{\theta}^{(j)} \gets \bm{\zeta}_{\mu} + \ln(1 + \exp(\bm{\zeta}_{\rho})) \odot \bm{z}^{(j)}$, $j = 1,2...N_z$, $\odot$ denotes element-wise product.\;
\State $L(\bm{\zeta}) \gets \frac{1}{N_z}\sum_{j=1}^{N_z}[\ln Q(\bm{\theta}^{(j)}; \bm{\zeta}) - \ln P(\bm{\theta}^{(j)}) - \ln P(\mathcal{D} | \bm{\theta}^{(j)})]$.\;
\State Update $\boldsymbol{\zeta}$ with gradient $\nabla_{\boldsymbol{\zeta}}L(\boldsymbol{\zeta})$ using Adam optimizer.\;
\EndFor
\State Sample $\{\boldsymbol{z}^{(j)}\}_{j=1}^{M}$ independently from $\mathcal{N}(\bm{0}, \bm{I}_{d_{\bm{\theta}}})$. \;
\State $\bm{\theta}^{(j)} \gets \bm{\zeta}_{\mu} + \ln(1 + \exp(\bm{\zeta}_{\rho})) \odot \bm{z}^{(j)}$, $j = 1,2...M$.\;
\State Calculate $\{ \tilde{u}(\boldsymbol{x},\boldsymbol{\theta}^{(j)})\}_{j=1}^M$ as samples of $u(\boldsymbol{x})$, similarly for other terms. \;

\end{algorithmic}
\end{algorithm}

\section{Results and Discussion}
\label{sec:results}
In this section we present a systematic comparison among the B-PINNs with different posterior sampling methods, i.e., HMC (B-PINN-HMC) and VI (B-PINN-VI), as well as the dropout \cite{zhang2019quantifying,gal2016dropout} for 1D function approximation, and 1D/2D forward/inverse PDE problems. 

In all the cases, we employ a neural networks with 2 hidden layers, each with width of 50, for B-PINNs. The prior for $\bm{\theta}$ is set as independent standard Gaussian distribution for each component. Such size of the neural network and the prior distribution are inherited from \cite{yao2019quality}. In the 1D case, the covariance function for $\tilde{u}$ is shown in Fig. A.\ref{fig:priorb} (\ref{sec:prior}). 
In HMC, the mass matrix is set to the identity matrix, i.e., $\bm{M} = \bm{I}$ \cite{yao2019quality}, the leapfrog step is set to $L = 50 \delta t$, the initial time step is $\delta t = 0.1$, the burn-in steps are set to $2,000$, and the total number of samples is $15,000$. 
In VI, the Adam optimizer is employed for training, and the total number of training steps is $N = 200,000$ with batch size $N_z = 5$. The hyperparameters for Adam optimizer are set as $l = 10^{-3}, ~\beta_1 = 0.9, ~\beta_2 = 0.999$. 
In the dropout method, we randomly drop a certain number of neurons with a predefined probability (i.e., dropout rate) at each training step \cite{gal2016dropout}. The number of the training steps is $200, 000$.  The hyperparameters for the Adam optimizer are set as $l = 10^{-3}, ~\beta_1 = 0.9, ~\beta_2 = 0.999$. To quantify the uncertainty, the strategy used in \cite{zhang2019quantifying} is also employed here, i.e., after finishing the training we run the forward propagation of DNNs $M$ times with the same dropout rate as in training, and then compute the mean and standard deviation based on the samples.
When estimating the means and standard deviations, we set the number of samples $M = 10,000$ for all the methods in each test case.

\subsection{Function regression}
In this section, we test the posterior sampling methods introduced above on a function regression task. In this task, the B-PINN is reduced to a BNN. The framework is the same as that for solving PDEs, except that our data set only involves the unknown function $u(x)$, thus the likelihood is $P(\mathcal{D}|\bm{\theta}) = P(\mathcal{D}_u|\bm{\theta}) $. The test function is expressed as
\begin{align}\label{eq:func}
    u(x) = \sin^3(6 x), ~x \in [-1, 1],
\end{align}
and we use 32 training points placed in $[-0.8, -0.2] \cup [0.2, 0.8]$, with observation noise $\epsilon_u \thicksim \mathcal{N}(0, 0.1^2)$.

\begin{figure}[H]
    \centering
    \subfigure[]{\label{fig:funca}
    \includegraphics[width=0.23\textwidth]{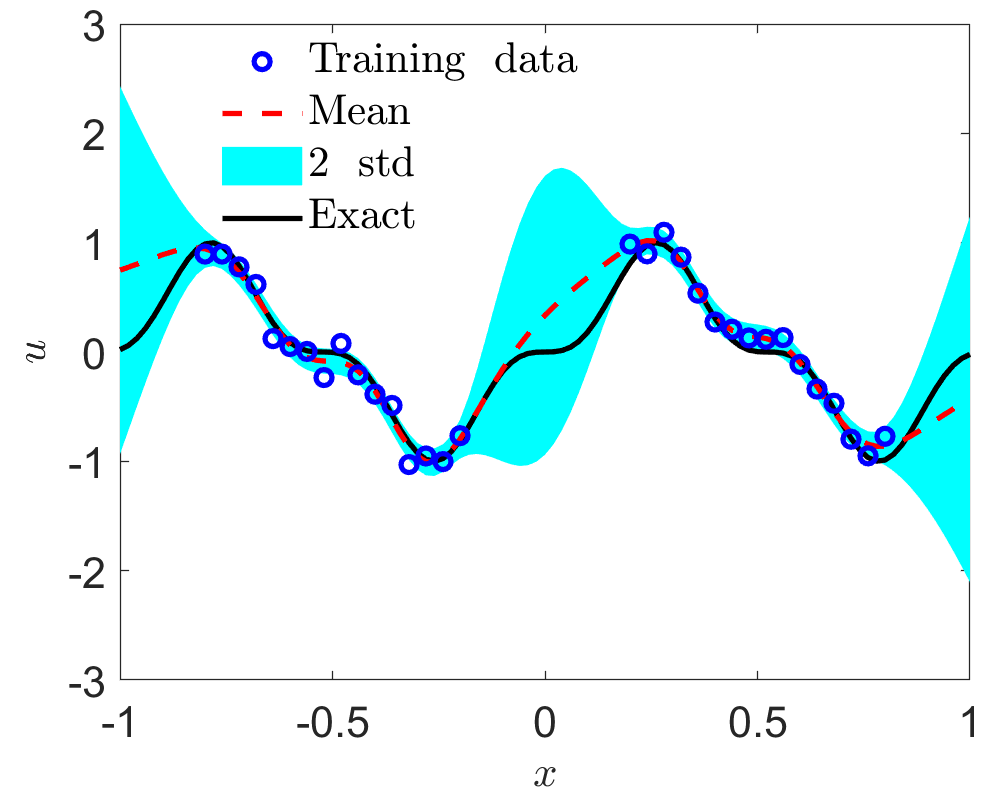}}
    \subfigure[]{\label{fig:funcb}
    \includegraphics[width=0.23\textwidth]{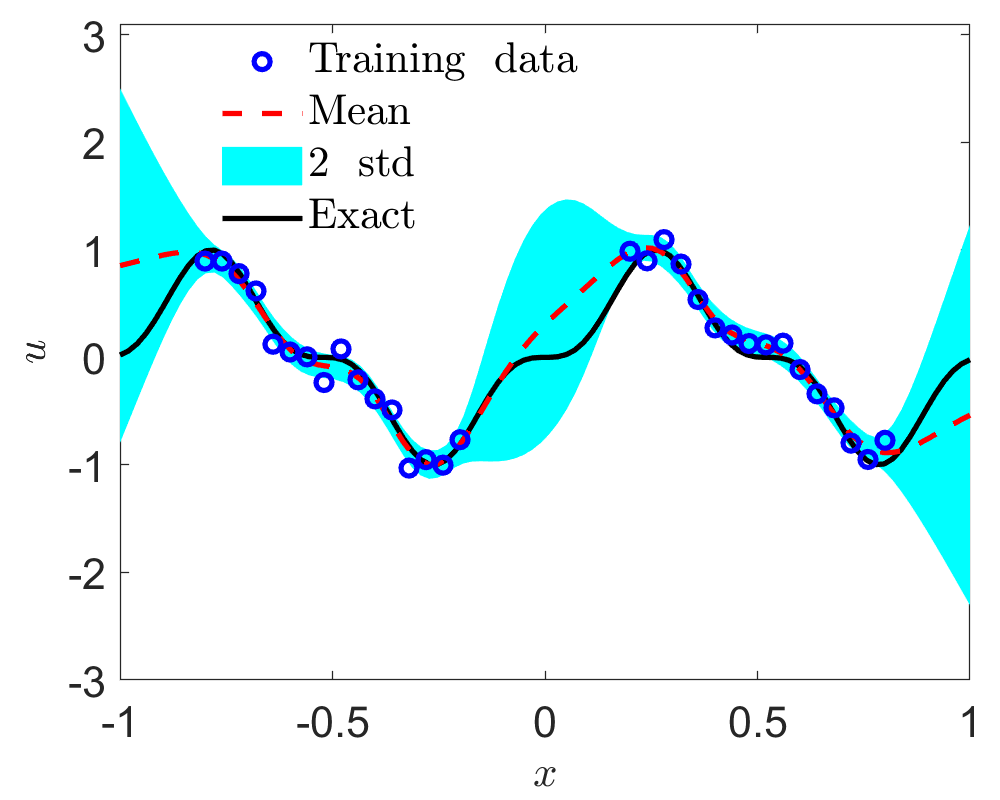}}
    \subfigure[]{\label{fig:funcc}
    \includegraphics[width=0.23\textwidth]{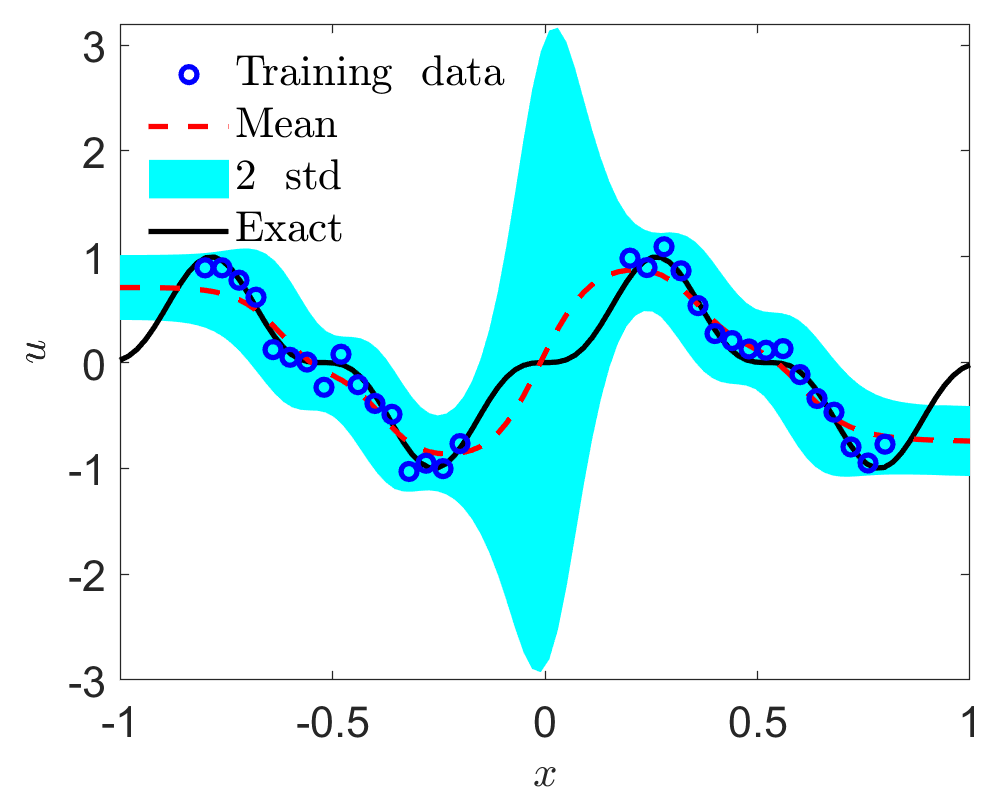}}\\
    \subfigure[]{\label{fig:funcd}
    \includegraphics[width=0.23\textwidth]{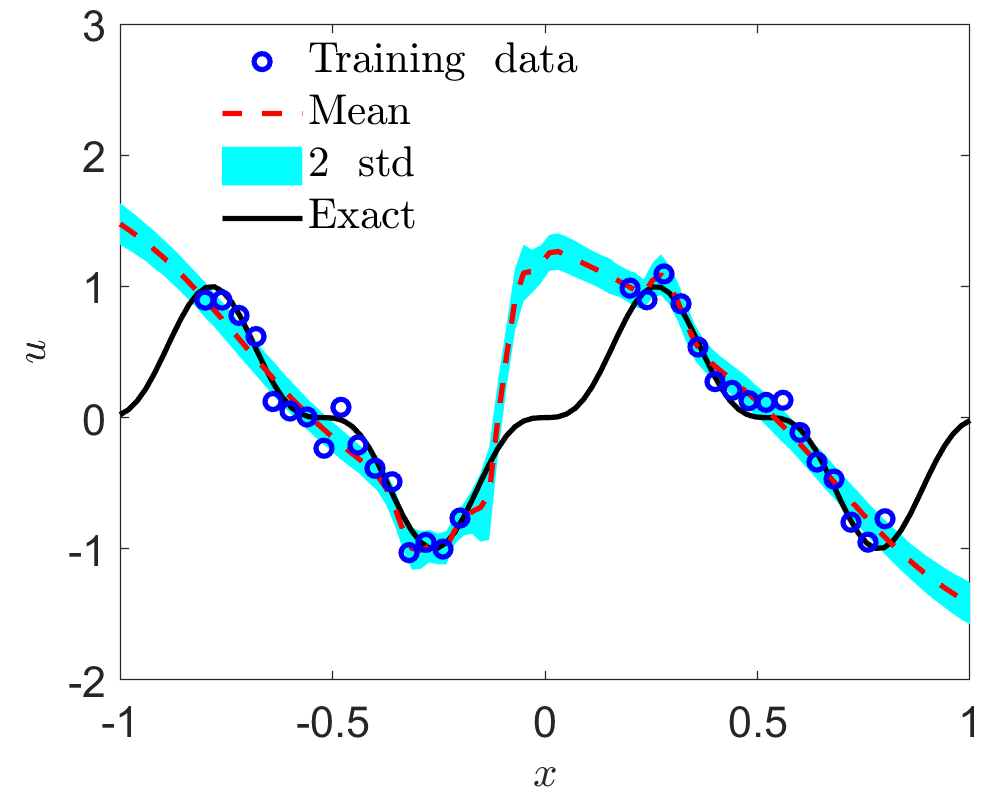}}
     \subfigure[]{\label{fig:funce}
    \includegraphics[width=0.23\textwidth]{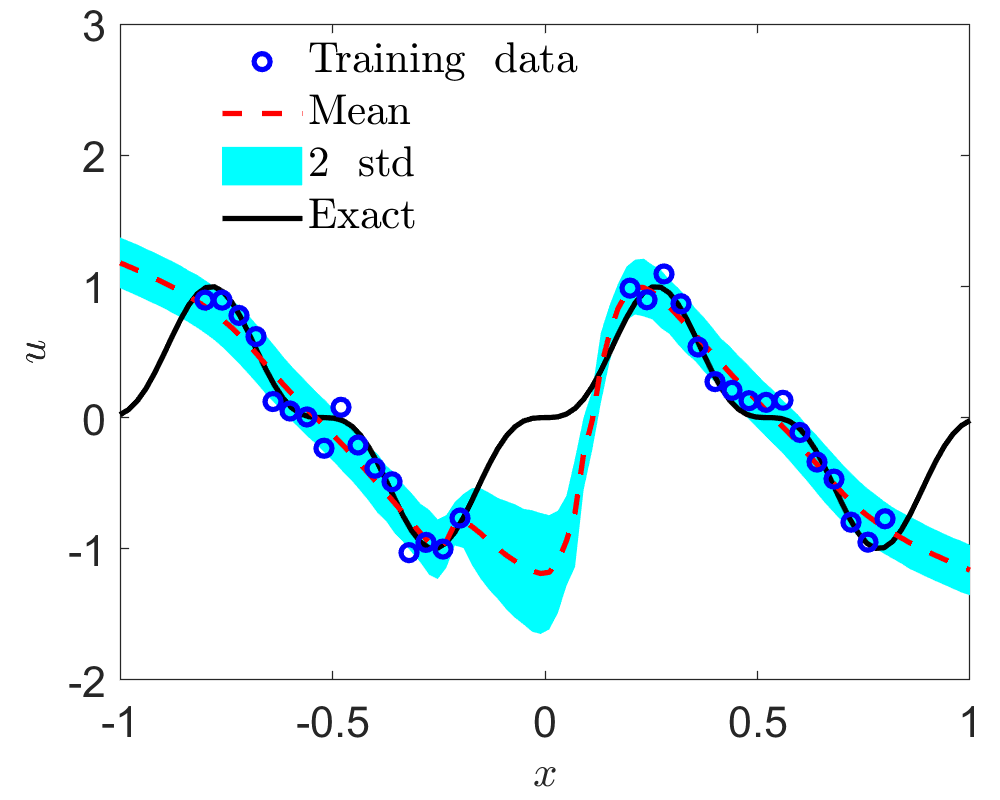}}
      \subfigure[]{\label{fig:funcf}
    \includegraphics[width=0.23\textwidth]{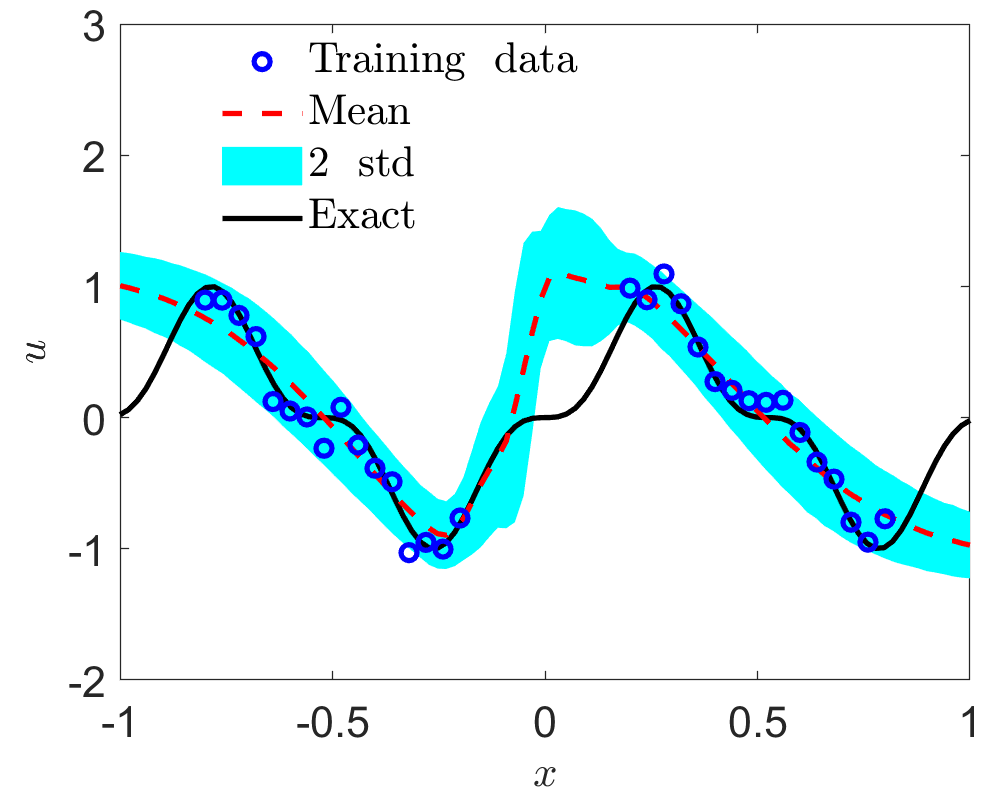}}
    \caption{Function approximation and comparison among  different  approaches. (a) BNN-GPR, (b) BNN-HMC, (c) BNN-VI, (d) Dropout with drop rate $1 \%$, (e) Dropout with drop rate $5 \%$, (f) Dropout with drop rate $20 \%$ (4 hidden layers with 100 neurons per layer).}
    \label{fig:func}
\end{figure}

We note that the prior for BNNs is a Gaussian process with zero mean as the width of each hidden layer goes to infinity \cite{neal2012bayesian,lee2017deep,pang2019neural}. In our case where the width is 50, we could see from Fig.~\ref{fig:density_u} that the prior at several single points matches well the Gaussian distribution. We thus view the prior as a Gaussian process approximately. Although the analytical expression for the kernel cannot be calculated, it could be estimated from independent samples of neural network functions ($100,000$ samples), as is illustrated in Fig. A.\ref{fig:priorb}. Therefore, GPR could be applied and provide reference solutions for the posterior estimations, denoted as BNN-GPR. Note that such reference solution is not analytical and the errors could come from the Gaussian process assumption for neural networks with finite width as well as the empirical estimation of the kernel. The results are illustrated in Fig. \ref{fig:func}. We can see from Figs. \ref{fig:funca}-\ref{fig:funcb} that (1) the predictive means are observed to be similar to the exact function $u(x)$, and the predicted standard deviations from these two methods are quite similar, and (2) the standard deviations (i.e., uncertainty)  becomes larger at the regions with fewer training data, i.e., $x \in [-1, -0.8] \cup [-0.2, 0.2] \cup [0.8, 1]$, which is reflected in the growing uncertainty due to lack of data. 

Note that the dimension of the unknown parameter $\bm{\theta}$ is $2701$ in our case using BNNs as prior. Despite the high dimensionality, HMC also provides posterior estimation (Fig. \ref{fig:funcb}) similar to the reference solution (Fig. \ref{fig:funca}) in this case, which shows its effectiveness in sampling from high dimensional distributions. As for the BNN-VI, the uncertainty at $x \in [-0.2, 0.2]$ is observed to be larger, while the uncertainty around $x \in [-1, -0.8] \cup [0.8, 1]$ is underestimated. Such results indicate that VI is not as accurate as HMC in posterior estimation, which could be attributed to the fact that the samples are actually drawn from $Q$ in Eq. \eqref{eqn:VI_Q}, which is limited to a family that could be too small to give a reasonable approximation of the posterior distribution.

Since both the architecture of the DNNs and the dropout rate are known to have strong effects on the certainty quantification, we test there different cases: (1) 2 hidden layers with 50 neurons and dropout rate 0.01, (2) 2 hidden layers with 50 neurons and dropout rate 0.05, and (3) 4 hidden layers with 100 neurons and dropout rate  0.2.  We use the same dropout rate at the prediction step as that used in the training process to quantify the uncertainty. As we can see from Figs. \ref{fig:funcd}-\ref{fig:funcf}, the uncertainties appear to be uniform for all the $x \in [-1, 1]$, which is not reasonable at all in this case. Furthermore, the predictive mean is quite different from the exact function $u(x)$ at the gap data zones, and the differences are far beyond the two standard deviations. To keep consistent, we will employ the same architecture of DNNs with the dropout as the BNNs, i.e., 2 hidden layers with 50 neurons in the following.

It is worth mentioning that while the priors for the BNNs with infinite width have been well studied \cite{neal2012bayesian,lee2017deep,pang2019neural}, the choice of priors for BNNs with finite width remains an open question. More discussion about the influence of architecture of the neural networks as well as the prior distributions of the parameters will be presented in \ref{sec:prior}.

\subsection{Forward PDE problems}
\label{sec:forward}

\subsubsection{1D Poisson equation}
\label{sec:linear}
 \begin{figure}[H]
    \centering
    \subfigure[]{\label{fig:lpdea}
   \includegraphics[width=0.8\textwidth]{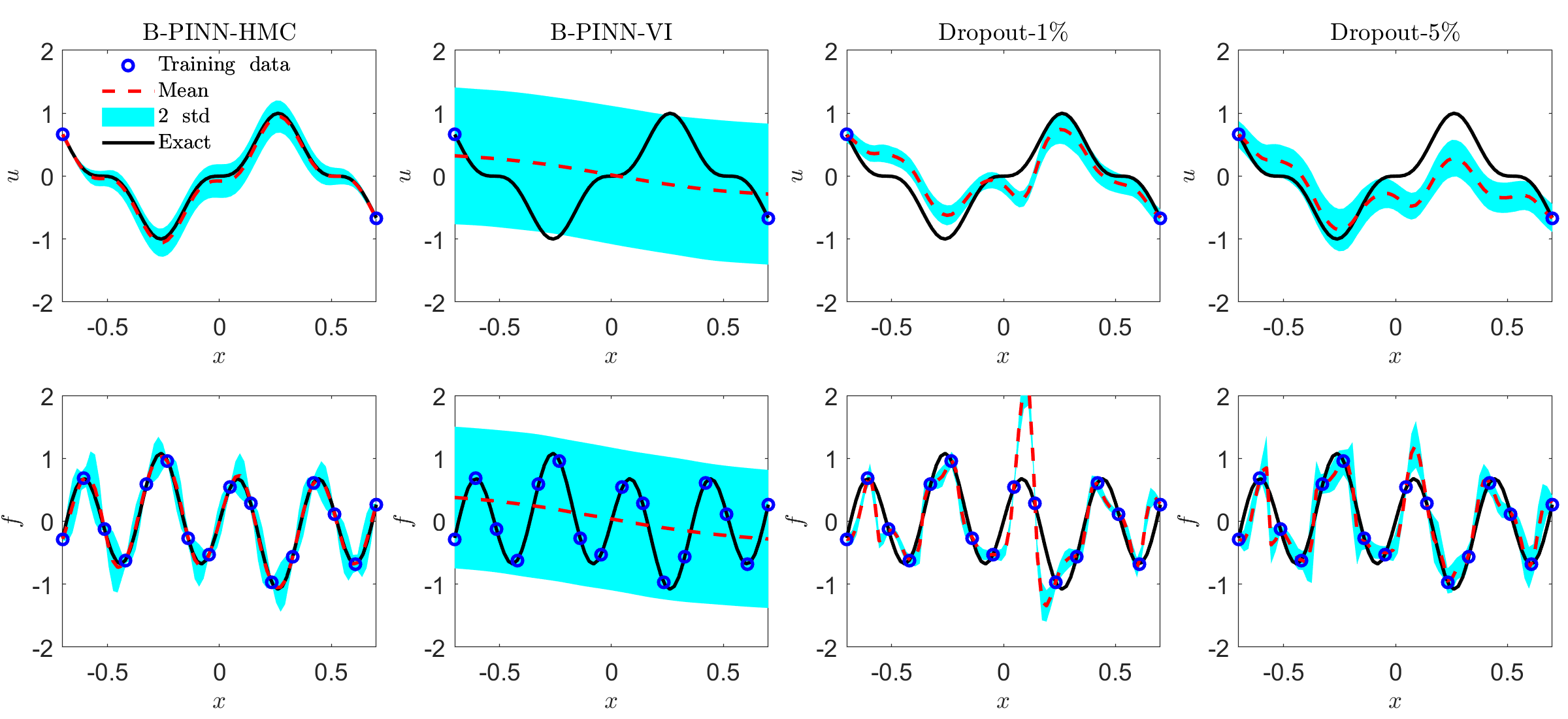}}
\subfigure[]{\label{fig:lpdeb}
   \includegraphics[width=0.8\textwidth]{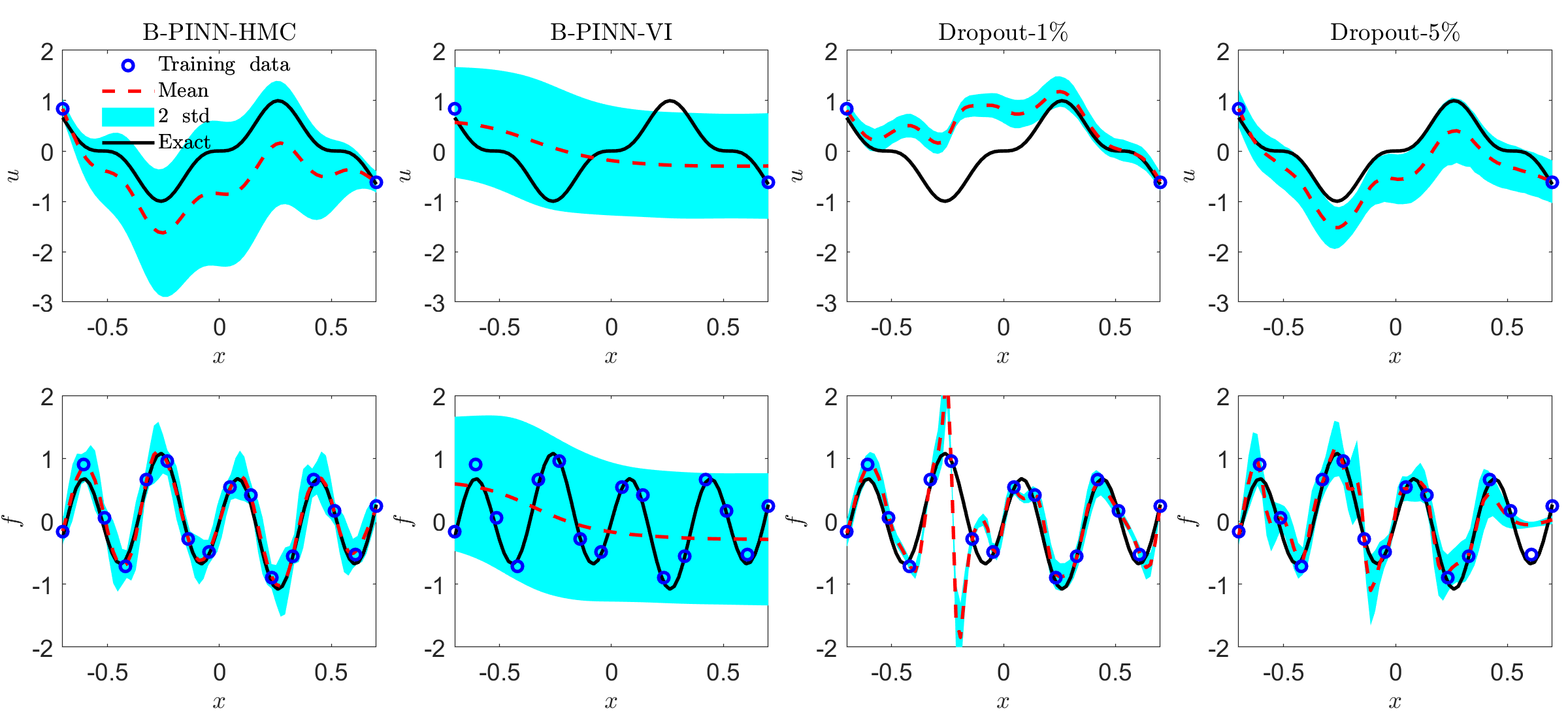}}
    \caption{1D linear Poisson equation - forward problem: predicted $u$ and $f$ from different methods with two data noise scales. (a) $\epsilon_f \thicksim \mathcal{N}(0, 0.01^2)$, $\epsilon_b \thicksim \mathcal{N}(0, 0.01^2)$. (b) $\epsilon_f \thicksim \mathcal{N}(0, 0.1^2)$, $\epsilon_b \thicksim \mathcal{N}(0, 0.1^2)$.}
    \label{fig:lpde}
\end{figure}

We consider the following linear Poisson equation:
\begin{align}\label{eq:linearpoisson}
    \lambda \partial^2_x u = f, x \in [-0.7, 0.7],
\end{align}
where $\lambda = 0.01$. The solution for $u$ is $u = \sin^3(6x)$, and $f$ can be derived from Eq. \eqref{eq:linearpoisson}. Here we assume that the exact expression of $f$ is unknown, but instead we have 16 sensors for $f$, which are equidistantly distributed in $x \in [-0.7, 0.7]$. Furthermore, we have two sensors at $x = -0.7$ and $0.7$ to provide the left/right Dirichlet boundary conditions for $u$. We assume that all the measurements from the sensors are noisy, and we consider the following two different cases: (1) $\epsilon_f \thicksim \mathcal{N}(0, 0.01^2)$, $\epsilon_b \thicksim \mathcal{N}(0, 0.01^2)$, and (2) $\epsilon_f \thicksim \mathcal{N}(0, 0.1^2)$, $\epsilon_b \thicksim \mathcal{N}(0, 0.1^2)$.  The results of B-PINN-HMC, B-PINN-VI and dropout, are illustrated in Fig. \ref{fig:lpde}.

We see that the B-PINN-HMC gives reasonable posterior estimation in that (1) the error between the means and the exact solution is mostly bounded by the two standard deviations, (2) the standard deviation increases with  increasing noise scale. However, B-PINN-VI completely failed to provide predictions to the solution $u$. Finally, the prediction given by dropout seems to match the data, but the predictive means for $u$ and $f$ differ from the exact solutions.  We also  note that the uncertainties provided by the dropout are not reasonable, e.g., (1) there is no significant differences between uncertainties for $u$ in the two cases with different noise scale, and (2) a large part of the exact solution does not lie in the two standard deviation confidence intervals.

\subsubsection{1D nonlinear Poisson equation}
\label{sec:1Dnonforward}

\begin{figure}[H]
    \centering
    \subfigure[]{\label{fig:nlpdea}
   \includegraphics[width=0.8\textwidth]{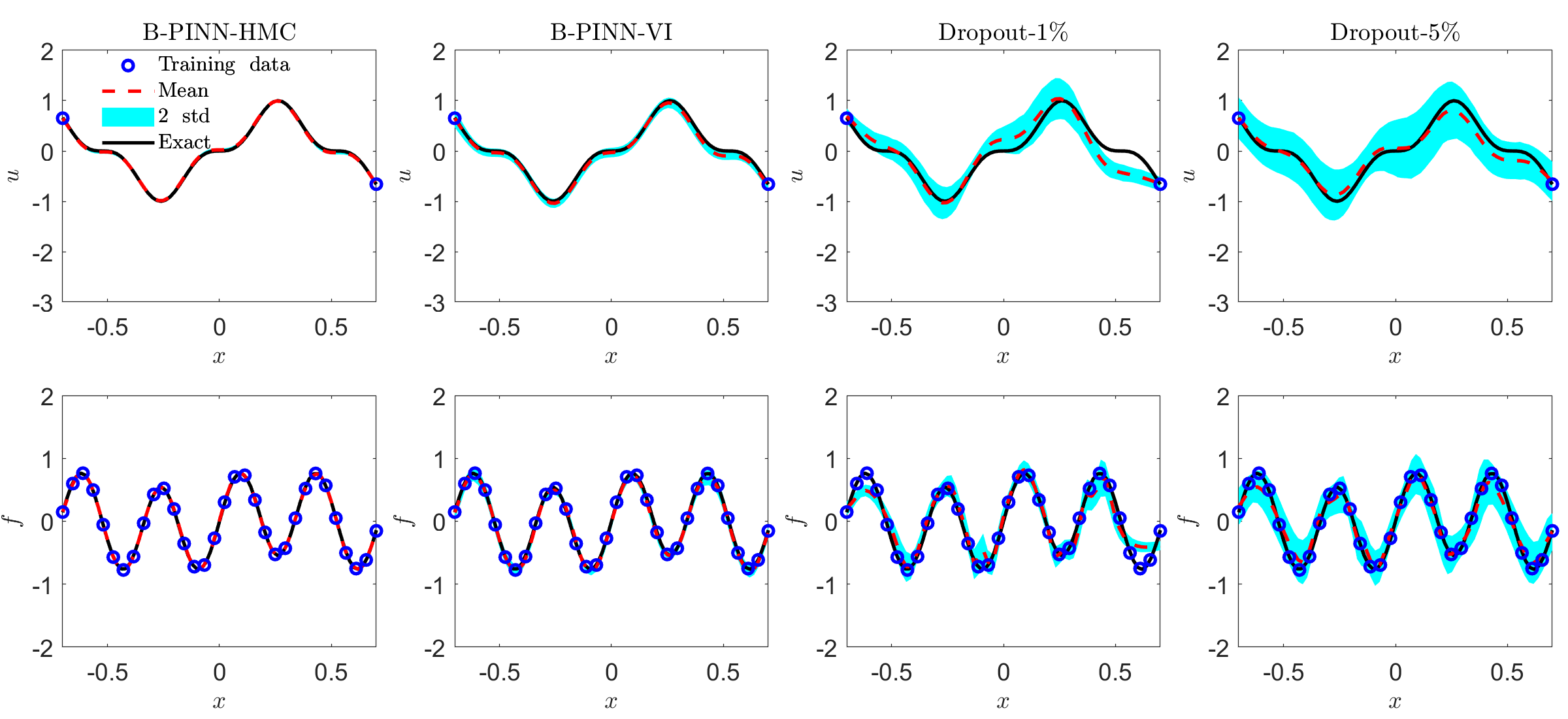}}
    \subfigure[]{\label{fig:nlpdeb}
   \includegraphics[width=0.8\textwidth]{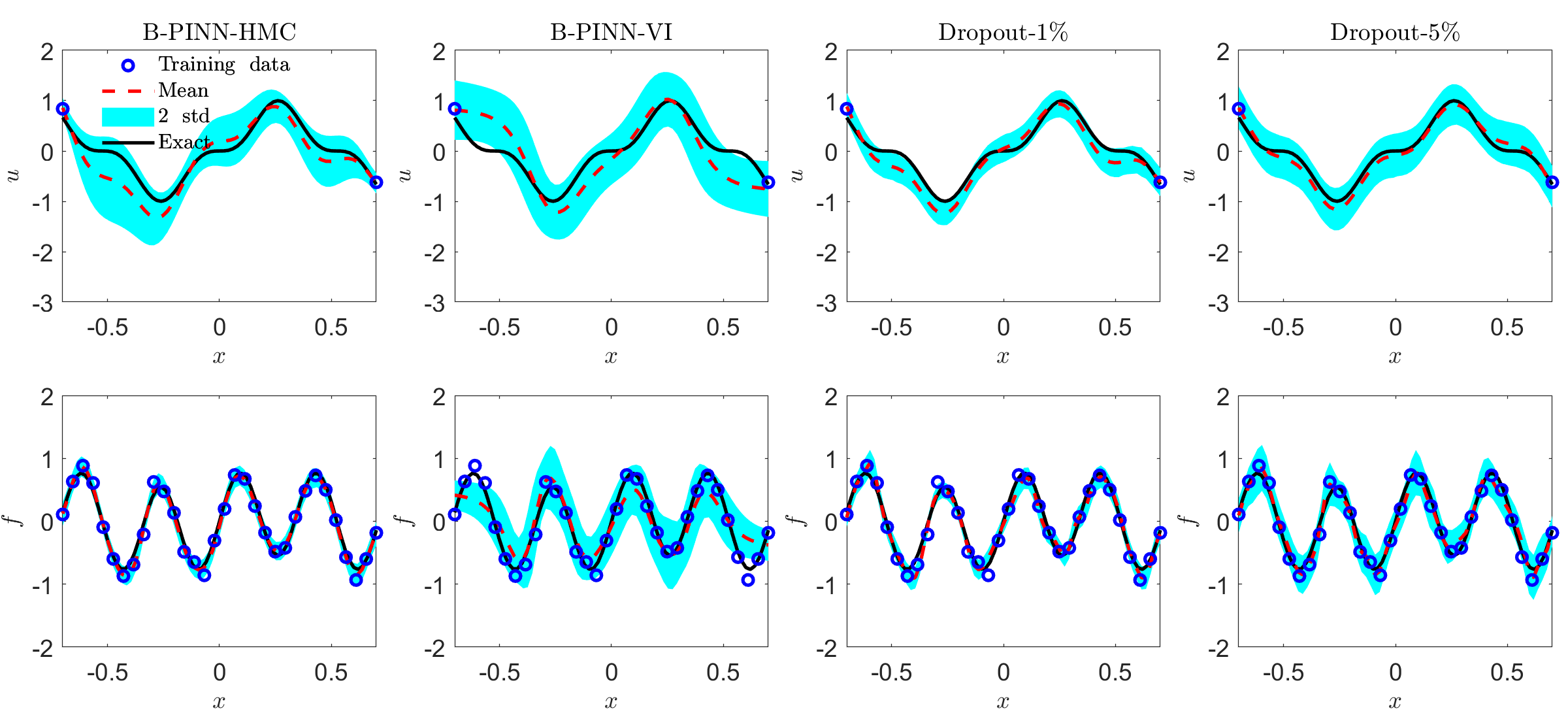}}
    \caption{1D nonlinear Poisson equation - forward problem: predicted $u$ and $f$ from different methods with two data noise scales. (a) $\epsilon_f \thicksim \mathcal{N}(0, 0.01^2)$, $\epsilon_b \thicksim \mathcal{N}(0, 0.01^2)$. (b) $\epsilon_f \thicksim \mathcal{N}(0, 0.1^2)$, $\epsilon_b \thicksim \mathcal{N}(0, 0.1^2)$.}
    \label{fig:nlpde}
\end{figure}

Here we consider the following 1D nonlinear PDE
\begin{align}\label{eq:nonlinear}
    \lambda \partial^2_x u + k \tanh(u) = f , x \in [-0.7, 0.7],
\end{align}
and we use the same solution for $u$ as the case in Sec. \ref{sec:linear}, i.e., $u = \sin^3(6x)$. In addition, $\lambda = 0.01$, and $k = 0.7$ is a constant, while $f$ can then be derived from Eq. \eqref{eq:nonlinear}. We assume that we have 32 sensors for $f$, which are equidistantly placed in $x \in [-0.7, 0.7]$. In addition, two sensors for $u$ are placed at $x = -0.7$ and $0.7$ to provide Dirichlet boundary conditions. Here, we also consider two different scales of Gaussian noise in the measurements, i.e., (1) $\epsilon_f \thicksim \mathcal{N}(0, 0.01^2)$, $\epsilon_b \thicksim \mathcal{N}(0, 0.01^2)$, and (2) $\epsilon_f \thicksim \mathcal{N}(0, 0.1^2)$, $\epsilon_b \thicksim \mathcal{N}(0, 0.1^2)$, which is the same as in Sec. \ref{sec:linear}.
The results of B-PINN-HMC, B-PINN-VI and dropout are illustrated in Fig. \ref{fig:nlpde}. 

The B-PINN-HMC provides good predictions for both $u$ and $f$, which are similar as the results in Sec. \ref{sec:linear}.  While able to give predicted means close to the exact solutions, B-PINN-VI and dropout can hardly give accurate uncertainty quantification for $u(x)$, e.g., (1) in the B-PINN-VI, the standard deviation for the case with noise scale 0.1 is observed to be large at the boundaries even though we have observations for the boundary conditions, (2) the noise scale has little influence on the predicted standard deviations in the two dropout cases.

\subsubsection{2D nonlinear Allen-Cahn equation}

\begin{figure}[H]
    \centering
    \subfigure[]{\label{fig:2dpdea}
    \includegraphics[width=0.8\textwidth]{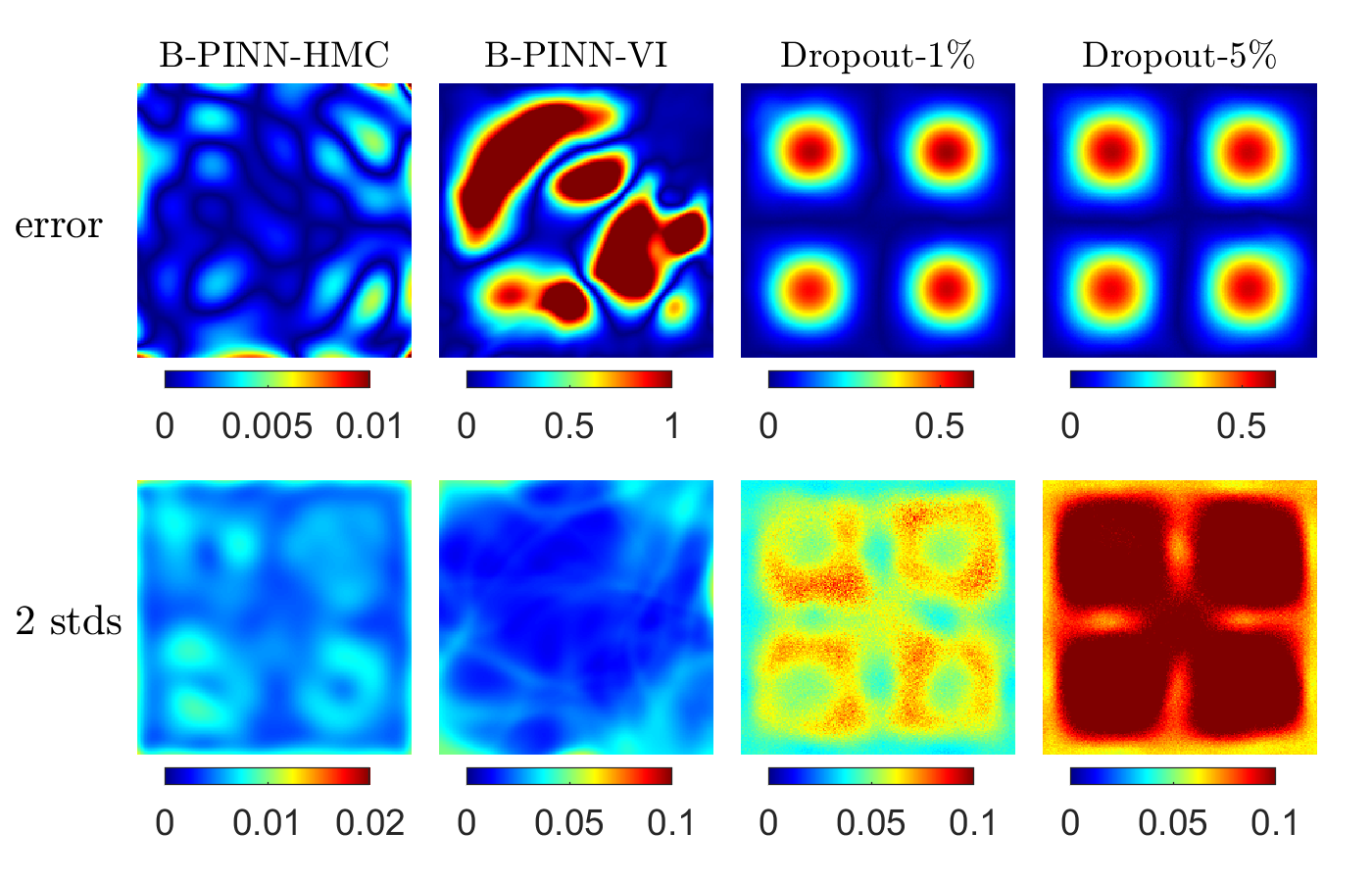}}
    \subfigure[]{\label{fig:2dpdeb}
    \includegraphics[width=0.8\textwidth]{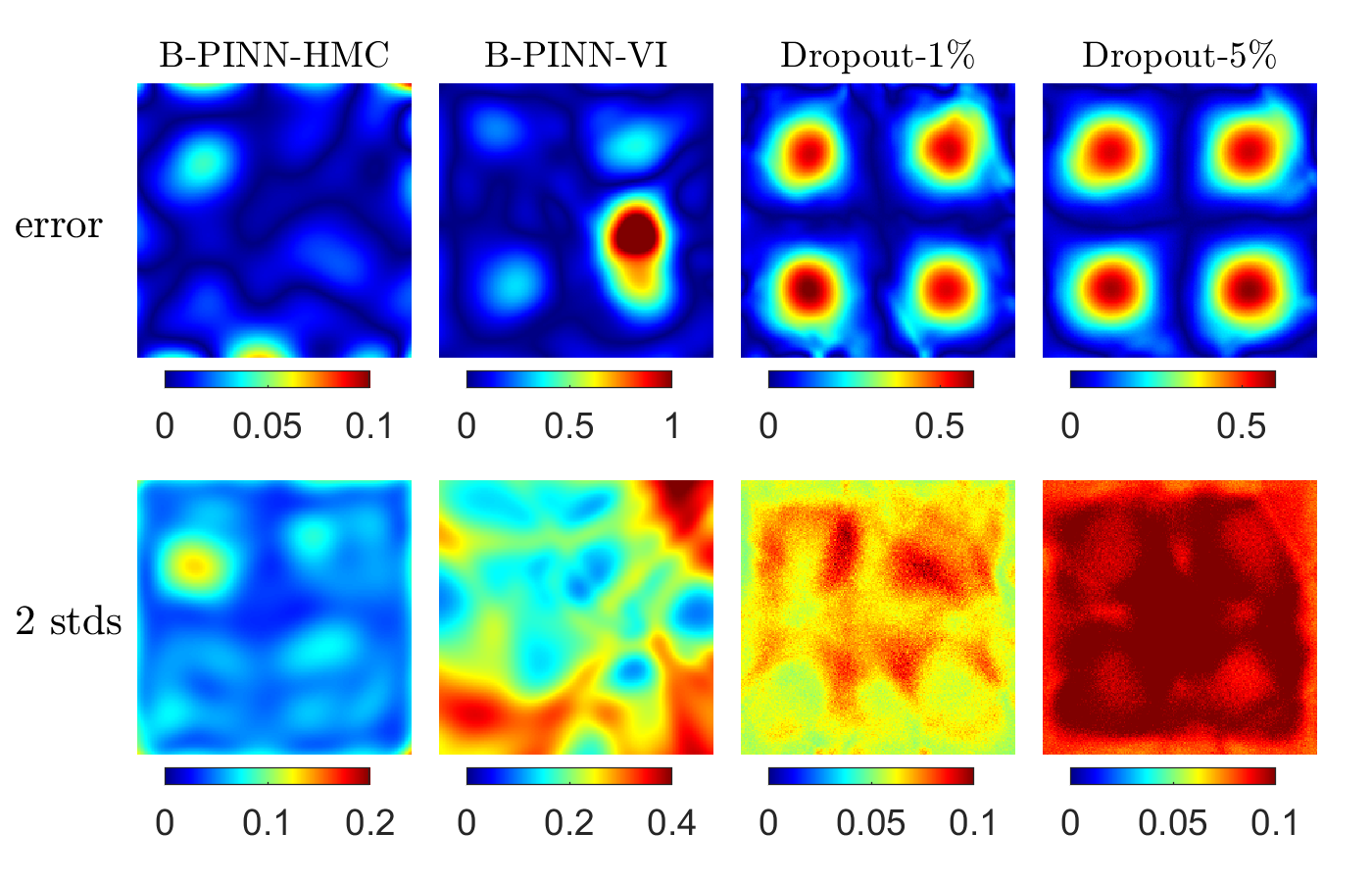}}
    \caption{
    2D Allen-Cahn equation - forward problem: Predicted errors and standard deviations for $u$ from different methods with two data noise scales. (a) $\epsilon_f \thicksim \mathcal{N}(0, 0.01^2)$, $\epsilon_b \thicksim \mathcal{N}(0, 0.01^2)$. (b) $\epsilon_f \thicksim \mathcal{N}(0, 0.1^2)$, $\epsilon_b \thicksim \mathcal{N}(0, 0.1^2)$.}
    \label{fig:2dpde}
\end{figure}

We further consider the following nonlinear Allen-Cahn equation which is a widely used model for multi-phase flows:
\begin{align}
    \lambda (\partial^2_x u + \partial^2_y u) +  u(u^2 - 1) = f, ~ x, y \in [-1, 1],
\end{align}
where $\lambda = 0.01$ represents the mobility, and $u$ is the order parameter, which denotes different phases.  Here, we employ the exact solution for $u = \sin(\pi x) \sin(\pi y)$. In addition, Dirichlet boundary conditions are imposed on all the boundaries. Similarly, we assume that we have 500 sensors for $f$, which are uniformly randomly distributed in the domain. In addition, we also have 25 equally distributed sensors for $u$ at each boundary. Here we also consider two different noise scales on all the measurements, i.e., (1) $\epsilon_f \thicksim \mathcal{N}(0, 0.01^2)$, $\epsilon_b \thicksim \mathcal{N}(0, 0.01^2)$, and (2) $\epsilon_f \thicksim \mathcal{N}(0, 0.1^2)$, $\epsilon_b \thicksim \mathcal{N}(0, 0.1^2)$.  The results of B-PINN-HMC, B-PINN-VI and dropouts, are illustrated in Fig.~\ref{fig:2dpde}.

Similarly, the B-PINN-HMC can provide predictive means close to the exact solutions, and the errors are mostly bounded by two standard deviations, which increases as the noise scale increases in the data. However, both the B-PINN-VI and the dropout with different drop rates fail to provide accurate means as well as uncertainties. Specifically, (1) the errors for the predicted $u$ from the B-PINN-VI and the two dropouts can be even larger than $50\%$ in part of the domain, (2) the errors for $u$ are not bounded by two standard deviations in the B-PINN-VI or in the two dropouts, and (3) the increase of the noise scale has little influence on the standard deviation when the dropout is employed.

\subsection{Inverse PDE problems}

\subsubsection{1D diffusion-reaction system with nonlinear source term}
\label{sec:inpde}
The PDE considered here is the same as Eq. \eqref{eq:nonlinear}. However, $k$ becomes an unknown parameter now. The objective here is to identify $k$ based on partial measurements of $f$ and $u$. 

We assume that we have 32 sensors for $f$, which are equidistantly placed in $x \in [-0.7, 0.7]$. In addition, two sensors for $u$ are placed at $x = -0.7$ and $0.7$ to provide Dirichlet boundary conditions. Apart from the boundary conditions, another 6 sensors for $u$ are placed in the interior of the domain to help identify $k$. We also assume that Gaussian noises are present for all the measurements. Two different scales of the noise are considered, i.e., (1) $\epsilon_f \thicksim \mathcal{N}(0, 0.01^2)$, $\epsilon_u \thicksim \mathcal{N}(0, 0.01^2)$, $\epsilon_b \thicksim \mathcal{N}(0, 0.01^2)$ and (2) $\epsilon_f \thicksim \mathcal{N}(0, 0.1^2)$, $\epsilon_u \thicksim \mathcal{N}(0, 0.1^2)$, $\epsilon_b \thicksim \mathcal{N}(0, 0.01^2)$. 

\begin{figure}[H]
    \centering
    \subfigure[]{\label{fig:1dipdea}
    \includegraphics[width=0.95\textwidth]{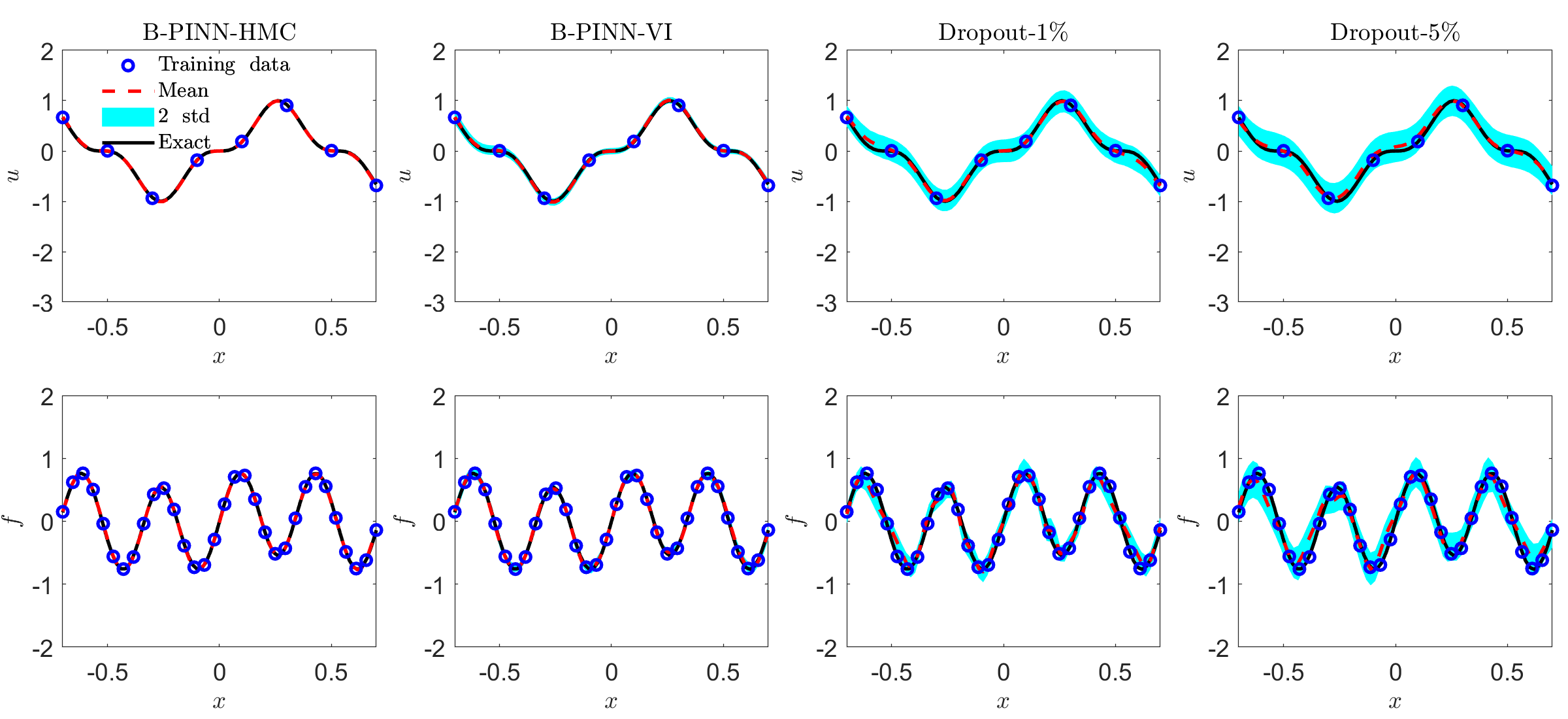}}
    \subfigure[]{\label{fig:1dipdeb}
    \includegraphics[width=0.95\textwidth]{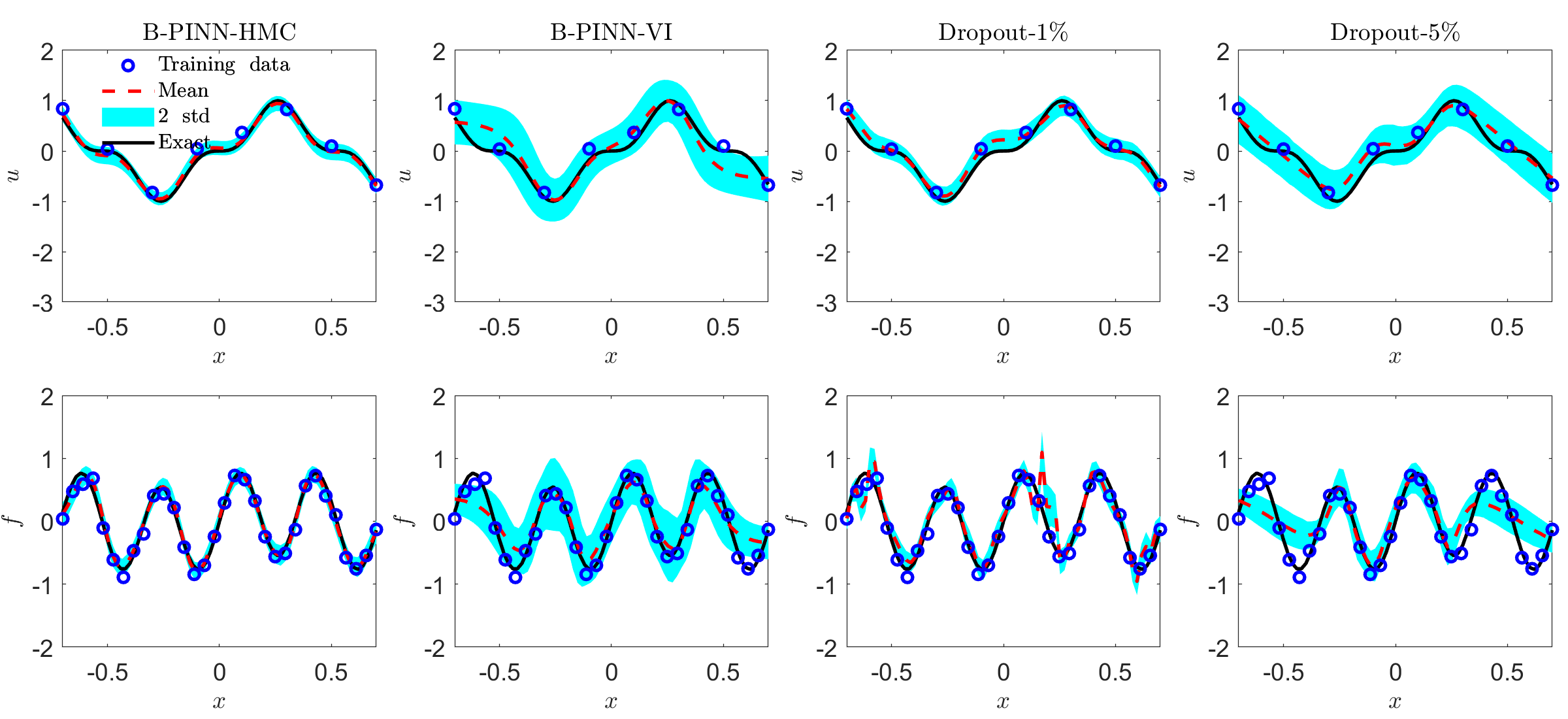}}
    \caption{1D diffusion-reaction system with nonlinear source term - inverse problem: predicted $u$ and $f$ from different methods with two data noise scales. (a) $\epsilon_f \thicksim \mathcal{N}(0, 0.01^2)$, $\epsilon_u \thicksim \mathcal{N}(0, 0.01^2)$, $\epsilon_b \thicksim \mathcal{N}(0, 0.01^2)$. (b) $\epsilon_f \thicksim \mathcal{N}(0, 0.1^2)$, $\epsilon_u \thicksim \mathcal{N}(0, 0.1^2)$, $\epsilon_b \thicksim \mathcal{N}(0, 0.01^2)$.}
    \label{fig:1dipde}
\end{figure}

The results of B-PINN-HMC, B-PINN-VI and dropouts are illustrated in Fig. \ref{fig:1dipde}. The predicted means of $u$ and $f$ from the B-PINN-HMC fit the exact functions well for both cases. The errors of $u$ and $f$ using B-PINN-VI and dropout are observed to be larger than those using B-PINN-HMC for the case with noise scale $0.1$.

The predicted values of $k$ from different methods are displayed in Table \ref{table:kpred}. The predicted means of $k$ for both cases using the B-PINN-HMC are quite accurate, with the error less than one standard deviation. Moreover, the standard deviation increases as the noise scale increases. The results show the effectiveness of the B-PINN-HMC in identifying the unknown parameter and quantifying the uncertainty arising from the scattered noisy data. The errors for B-PINN-VI are larger than B-PINN-HMC, although bounded by two standard deviations.  As for the dropout, we use $k$ from the last $10,000$ training steps as samples to calculate the mean and standard deviation, which is different from the posterior sampling used above. We observe that: (1) in both cases of noise scale, the errors of the predicted means with both dropout rates are larger than those from B-PINN-HMC, (2) the error increases as we increase the dropout rates, (3) for the case with dropout rate $5\%$, the standard deviation decreases as we increase the noise scale, which is not reasonable.

\begin{table}[H]
\centering
{\footnotesize
\begin{tabular}{cc|cccc}
\hline \hline
  \multicolumn{2}{c|}{ Noise scale}  & { B-PINN-HMC} & { B-PINN-VI} & { Dropout-1\%} & {Dropout-5\%} \\
  \hline
  \multirow{2}{*}{0.01} &Mean  & 0.705 & 0.708 & 0.714 & 0.669\\
  & Std   &  {\footnotesize $5.75 \times 10^{-3}$} & {\footnotesize $4.01 \times 10^{-3}$} & {\footnotesize $4.38 \times 10^{-3}$} & {\footnotesize $2.02 \times 10^{-2}$}\\
  \multirow{2}{*}{0.1} &Mean   & 0.665 & 0.775 & 0.746 & 0.633 \\
  & Std  & {\footnotesize $5.63 \times 10^{-2}$} & {\footnotesize $3.58 \times 10^{-2}$} & {\footnotesize $6.508 \times 10^{-3}$}  & {\footnotesize $6.45 \times 10^{-3}$}\\
  \hline \hline
\end{tabular}
}
\caption{1D diffusion-reaction system with nonlinear source term: Predicted mean and standard deviation for $k$ using different uncertainty quantification methods. The exact solution for $k$ is $0.7$. }
\label{table:kpred}
\end{table}

\subsubsection{2D nonlinear diffusion-reaction system}
\begin{figure}
\centering
\subfigure[]{
\includegraphics[width=0.45\textwidth]{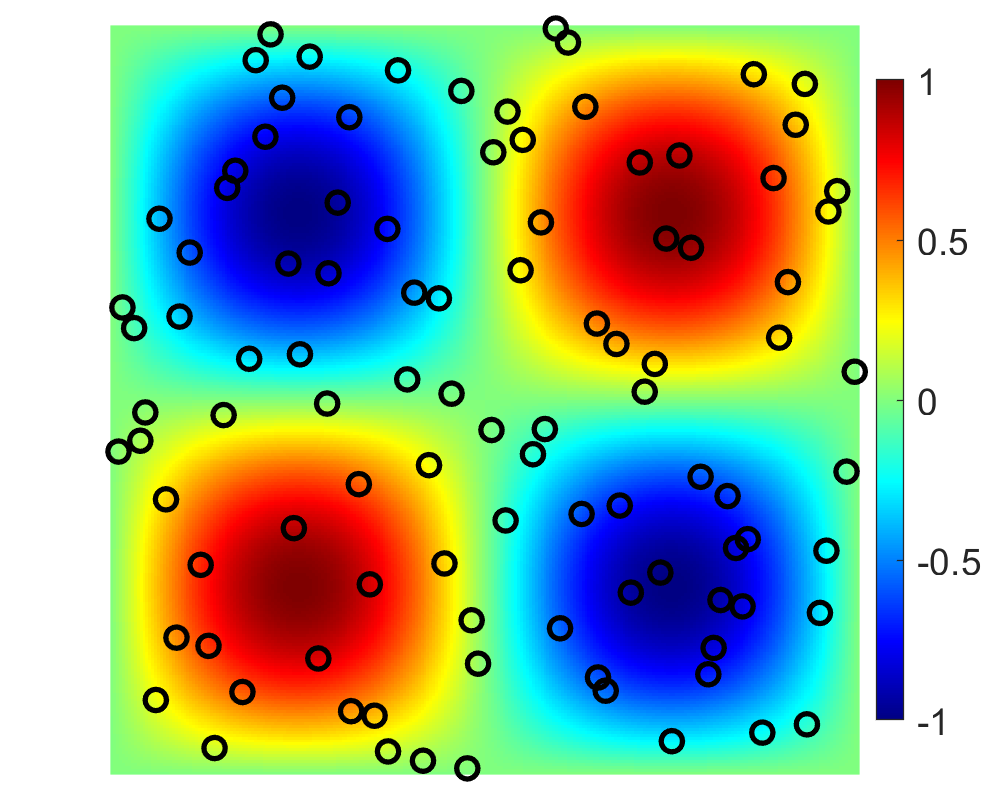}}
\subfigure[]{
\includegraphics[width=0.45\textwidth]{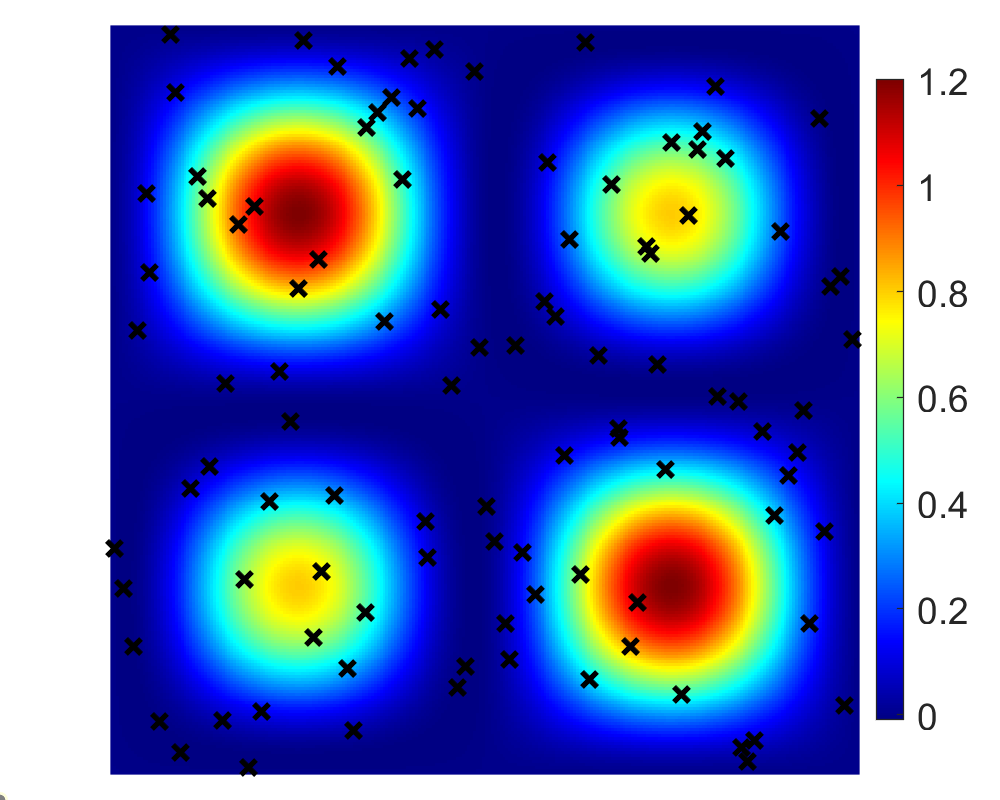}}
\caption{2D nonlinear diffusion-reaction system: Training data for $u$  and $f$. (a) Distribution of $u$. Black circle: training sample for $u$, (b) Distribution of $f$. Black cross: training samples for $f$.}
\label{fig:2d_train}
\end{figure}

We consider the following PDE here 
\begin{align}
    \lambda (\partial^2_x u + \partial^2_y u) + k u^2 =  f, ~ x, y \in [-1, 1],
\end{align}
where $\lambda = 0.01$ is the diffusion coefficient, $k$ represents the reaction rate, which is a constant, and $f$ denotes the source term. Here we assume that the exact value for $k$ is unknown, and we only have sensors for $u$ and $f$.  Specifically, we have 100 sensors which are randomly sampled in the physical domain (Fig. \ref{fig:2d_train}) for $u$ and $f$. we also have 25 equally distributed sensors for $u$ at each boundary for the Dirichlet boundary condition. Similarly, all the measurements are noisy and two noise scales are considered here, i.e., (1) $\epsilon_f \thicksim \mathcal{N}(0, 0.01^2)$, $\epsilon_u \thicksim \mathcal{N}(0, 0.01^2)$, $\epsilon_b \thicksim \mathcal{N}(0, 0.01^2)$, and (2) $\epsilon_f \thicksim \mathcal{N}(0, 0.1^2)$, $\epsilon_u \thicksim \mathcal{N}(0, 0.1^2)$, $\epsilon_b \thicksim \mathcal{N}(0, 0.01^2)$.  We then aim to estimate the reaction rate $k$ given the measurements of $u$ and $f$.

As we can see in Table \ref{table:2d_kpred}, for both cases with noise scale $0.1$ and $0.01$, the predictive means using the B-PINN-HMC are quite accurate with errors less than $5\%$. Also, the standard deviations, which represent the uncertainties, are in the same order of magnitude as the errors of predictive means. The uncertainty decreases as the scale of noise in data decreases. The results show the effectiveness of these two models in identifying the unknown parameter and quantifying the uncertainties arising from the noise in data. As for the B-PINN-VI, the relative errors between the predicted mean value of $k$ from and the exact $k$ are greater than $10\%$ for both cases. 
With regards to the dropout, the means and the standard deviations for $k$ are calculated in the same way as in Sec. \ref{sec:inpde}. The predicted means for $k$ from the dropout with dropout rate 0.01 are quite close to the exact $k$. As the dropout rate increases to 0.05, the errors become about $17\%$ for both cases. The above results show that the dropout rate has a strong effect on the predictive accuracy. However, there is no theory on the choice of the optimal dropout, which clearly restricts its usefulness in applications.  In addition, the influence of the noise scales on the standard deviations is not significant, indicating dropout is not suitable for quantifying uncertainty from noisy data.

\begin{table}
\centering
{\footnotesize
\begin{tabular}{cc|ccccc}
\hline \hline
  \multicolumn{2}{c|}{ Noise scale}  & { B-PINN-HMC} & { B-PINN-VI} & { Dropout-$1\%$} & { Dropout-$5\%$} \\
  \hline
  \multirow{2}{*}{0.01} &Mean  & 1.003 & 0.895 & 1.050 & 1.168 \\
  & Std   &   {\footnotesize $5.75 \times 10^{-3}$} & {\footnotesize $2.83 \times 10^{-3}$} & {\footnotesize $2.00 \times 10^{-3}$} & {\footnotesize $3.04 \times 10^{-3}$}\\
  \multirow{2}{*}{0.1} &Mean   & 0.978 & 1.116 & 1.020 & 1.169 \\
  & Std    & {\footnotesize $4.98 \times 10^{-2}$} & {\footnotesize $3.45 \times 10^{-2}$} & {\footnotesize $4.21 \times 10^{-3}$}  & {\footnotesize $4.15 \times 10^{-3}$}\\
  \hline \hline
\end{tabular}
}
\caption{2D nonlinear diffusion-reaction system: Predicted mean and standard deviation for the reaction rate $k$ using different certainty-induced methods. $k = 1$ is the exact solution.}
\label{table:2d_kpred}
\end{table}

\section{Comparison with PINNs}
\label{sec:pinn}

\begin{figure}[H]
    \centering
    \subfigure[]{\label{fig:pinninversea}
    \includegraphics[width=0.45\textwidth]{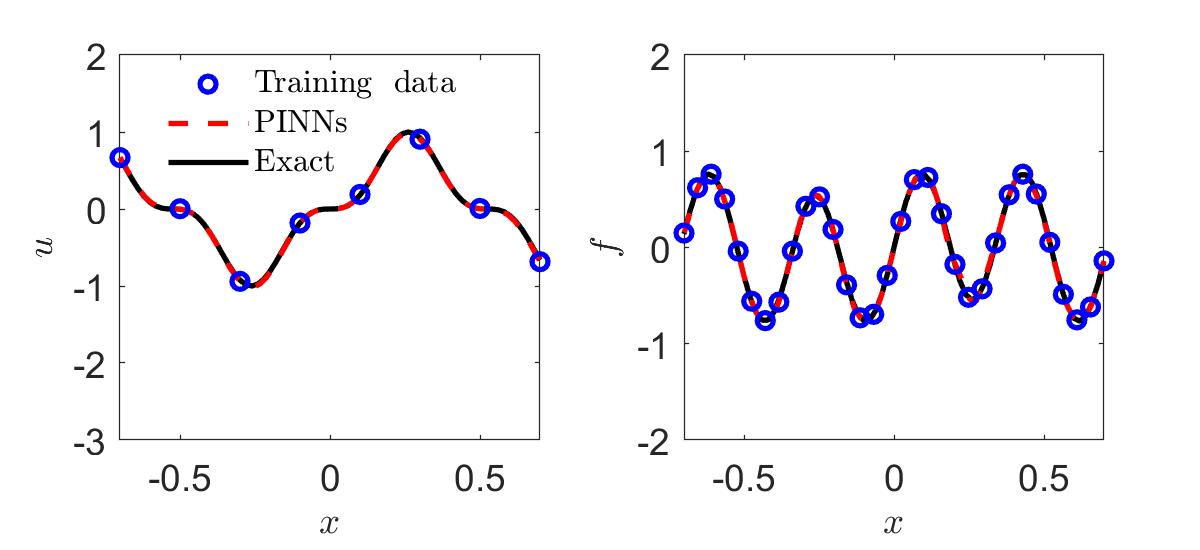}}
    \subfigure[]{\label{fig:pinninverseb}
    \includegraphics[width=0.45\textwidth]{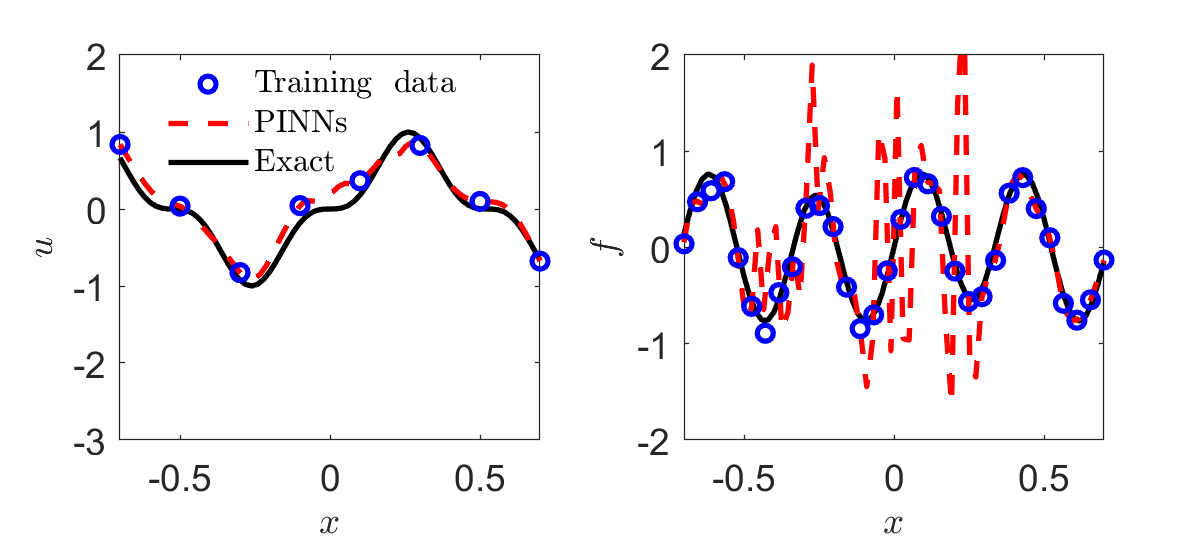}}
    \caption{1D diffusion-reaction system with nonlinear source term (PINNs): Predicted $u$ and $f$ with two data noise scales.   (a) $\epsilon_f \thicksim \mathcal{N}(0, 0.01^2)$, $\epsilon_u \thicksim \mathcal{N}(0, 0.01^2)$, $\epsilon_b \thicksim \mathcal{N}(0, 0.01^2)$. (b) $\epsilon_f \thicksim \mathcal{N}(0, 0.1^2)$, $\epsilon_u \thicksim \mathcal{N}(0, 0.01^2)$, $\epsilon_b \thicksim \mathcal{N}(0, 0.1^2)$.
    }
    \label{fig:pinninverse}
\end{figure}

In this section, we will conduct a comparison between the B-PINN-HMC and PINN for the 1D inverse problem in Sec. \ref{sec:inpde}. We employ the Adam optimizer with $l = 10^{-3},  ~ \beta_1 = 0.9, ~ \beta_2 = 0.999$ to train the PINN, with the number of the training steps set as $200, 000$. The results of the PINN are shown in Fig.~\ref{fig:pinninverse}. Note that the PINNs cannot quantify uncertainties of the predictive results.

As shown in Fig.~\ref{fig:pinninverse}, the predicted $u$ and $f$ could fit all the training points. In the cases where the noise scale is as small as $0.01$, the predictive $u$ and $f$ agree well with the exact solutions. However, as the noise scale increases to 0.1, significant overfitting is observed in PINNs. In addition, PINNs predict $k$ to be $0.705$ and $0.591$ for the noise lebel at $0.01$ and $0.1$, respectively, while the reference exact solution is $0.7$. Comparing with the results of B-PINN-HMC in Table~\ref{table:kpred}, we conclude that PINNs can provide prediction with similar accuracy as the B-PINN-HMC for the case with small noise in data, while B-PINN-HMC shows significant advantange in accuracy over  PINNs for the case with large noise.

Now, we conduct a brief comparison on the computational cost between PINN and B-PINN-HMC based on the inverse problem. We run both the PINN and B-PINN-HMC codes on two CPUs (Intel Xeon E5-2643). For the PINN, the computational time is about 10 minutes, while it takes about 20 minutes for the B-PINN-HMC. Despite this relatively small increase for B-PINN verse PINN for this small problem, we expect that when we scale up the data size and neural network size the difference in cost will increase accordingly. Considering the accuracy as well as the reliable uncertainty provided, the B-PINN-HMC may be a better approach than the PINNs for scenarios with large noise.

\section{Comparison with the truncated Karhunen-Lo\`eve expansion}
\label{sec:kl}
So far we have shown the effectiveness of B-PINNs in solving PDE problems. As we know, a neural network is extremely overparametrized. Hence, we want to investigate if we can we use other models with less parameters for our surrogate model in the Bayesian framework. For example, we consider the Karhunen-Lo\`eve expansion, a widely used representation for a stochastic process in the following study.

\subsection{Truncated Karhunen-Lo\`eve expansion}
Assume $u(\boldsymbol{x})$ is a stochastic process with mean $\mu(\boldsymbol{x})$ and covariance function (also called  ``kernel'') $k(\boldsymbol{x}, \boldsymbol{x}')$, then the KL expansion of $u$ is
\begin{equation}
\begin{aligned}
u(\boldsymbol{x}) = \mu(\boldsymbol{x}) + \sum_{i=1}^\infty\sqrt{\alpha_i} \psi_i(\boldsymbol{x}) \theta_i,
\end{aligned}
\end{equation}
where $\psi_i$ are the orthogonal eigenfunctions, $\alpha_i$ are the corresponding eigenvalues of the kernel, and $\theta_i$ are mutually uncorrelated random variables. In practice, we could truncate the expansion to $n$ terms as our surrogate model for $u$:
\begin{equation}
\begin{aligned}
\tilde{u}(\boldsymbol{x}; \boldsymbol{\theta}) = \mu(\boldsymbol{x}) + \sum_{i=1}^n\sqrt{\alpha_i} \psi_i(\boldsymbol{x}) \theta_i,
\end{aligned}
\end{equation}
where $\boldsymbol{\theta} = (\theta_1, \theta_2...\theta_n)$ is the parameter in the surrogate model whose prior distribution is given by the KL expansion.

One of the main differences between the truncated Karhunen-Lo\`eve expansion and neural networks as surrogate models is the number of parameters.  For example, in this paper, the neural network used in 1D problems has 2701 parameters. As a comparison, the truncated Karhunen-Lo\`eve expansion used in Sec.~\ref{sec:KLresults} only has 20 parameters. The small number of parameters makes it possible to use another approach to sample from the posterior, namely the deep normalizing flow (DNF) models.

In general, using  DNF to sample from a target distribution $\nu$ consists of the following three steps \cite{yang2019potential}:

\begin{enumerate}
    \item Define a bijective transformation $G: R^{d_{\boldsymbol{\theta}}} \rightarrow R^{d_{\boldsymbol{\theta}}}$ and prescribe an input distribution $\mu_I$ of the dimension $R^{d_{\boldsymbol{\theta}}}$. Usually, the bijective transformation is parameterized by deep neural networks and the input distribution can be a standard multivariate Gaussian distribution. 
    \item Note that the bijective transformation $G$ will map the input distribution to an output distribution $\mu_O = G_\# \mu_I$. We then train the parameters in $G$ to minimize $F(\mu_O, \nu)$, where $F$ is a functional that measures the difference between two distributions. Ideally, $\mu_O$ and $\nu$ will be sufficiently close to each other after this procedure.
    \item Finally, we sample from the input distribution $\mu_I$, denoted as $\{\boldsymbol{z}^{(j)}\}_{j=1}^M$. Then $\{G(\boldsymbol{z}^{(j)})\}_{j=1}^M$ as samples of $\mu_O$ can be used  to approximate the statistics of $\nu$.
\end{enumerate}

We leave the details of the DNF in \ref{sec:DNF}.

\subsection{Results and Comparisons}
\label{sec:KLresults}
In this section we apply the truncated Karhunen-Lo\`eve expansion to solve the forward and inverse nonlinear PDE problems as described in Sec.~\ref{sec:1Dnonforward} and Sec.~\ref{sec:inpde}.  In particular, we consider the Gaussian process of zero mean and exponential kernel 
\begin{equation}
    k(x,x') = \exp(-\frac{|x-x'|}{0.25}), \quad x\in [-1,1],
\end{equation}
and use the first 20 terms of the KL expansion as our surrogate model for $u$, which retains about $92\%$ of the energy. For this case, the eigenvalues and eigenfunctions in the KL expansion are solved analytically, and the prior for the unknown parameters is the product of independent standard Gaussian distributions. We refer the readers to example 4.1 in Chapter 4 in \cite{xiu2010numerical} for details.

\begin{figure}
    \centering
    \subfigure[]{
    \includegraphics[width=0.45\textwidth]{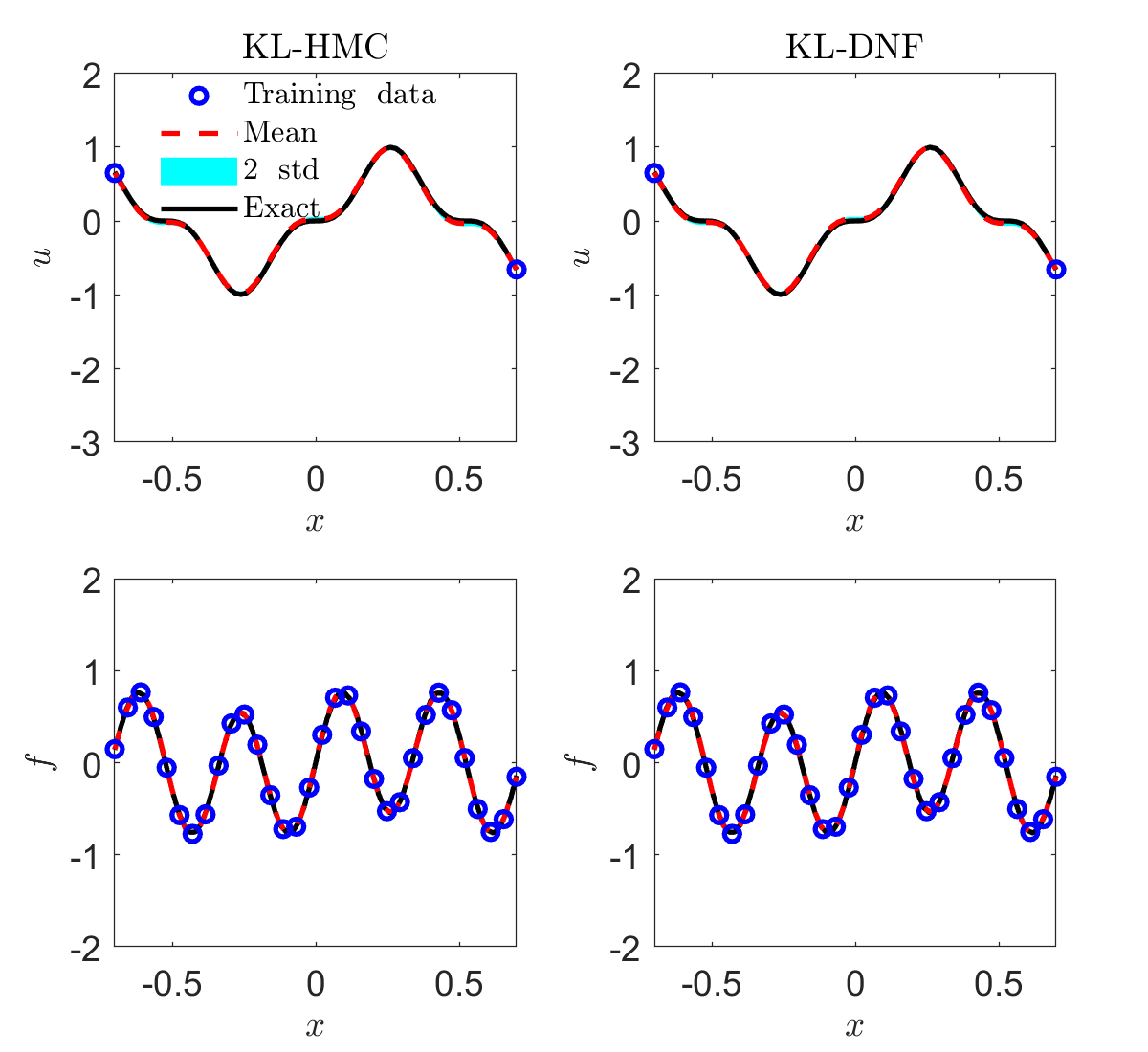}}
    \subfigure[]{
    \includegraphics[width=0.45\textwidth]{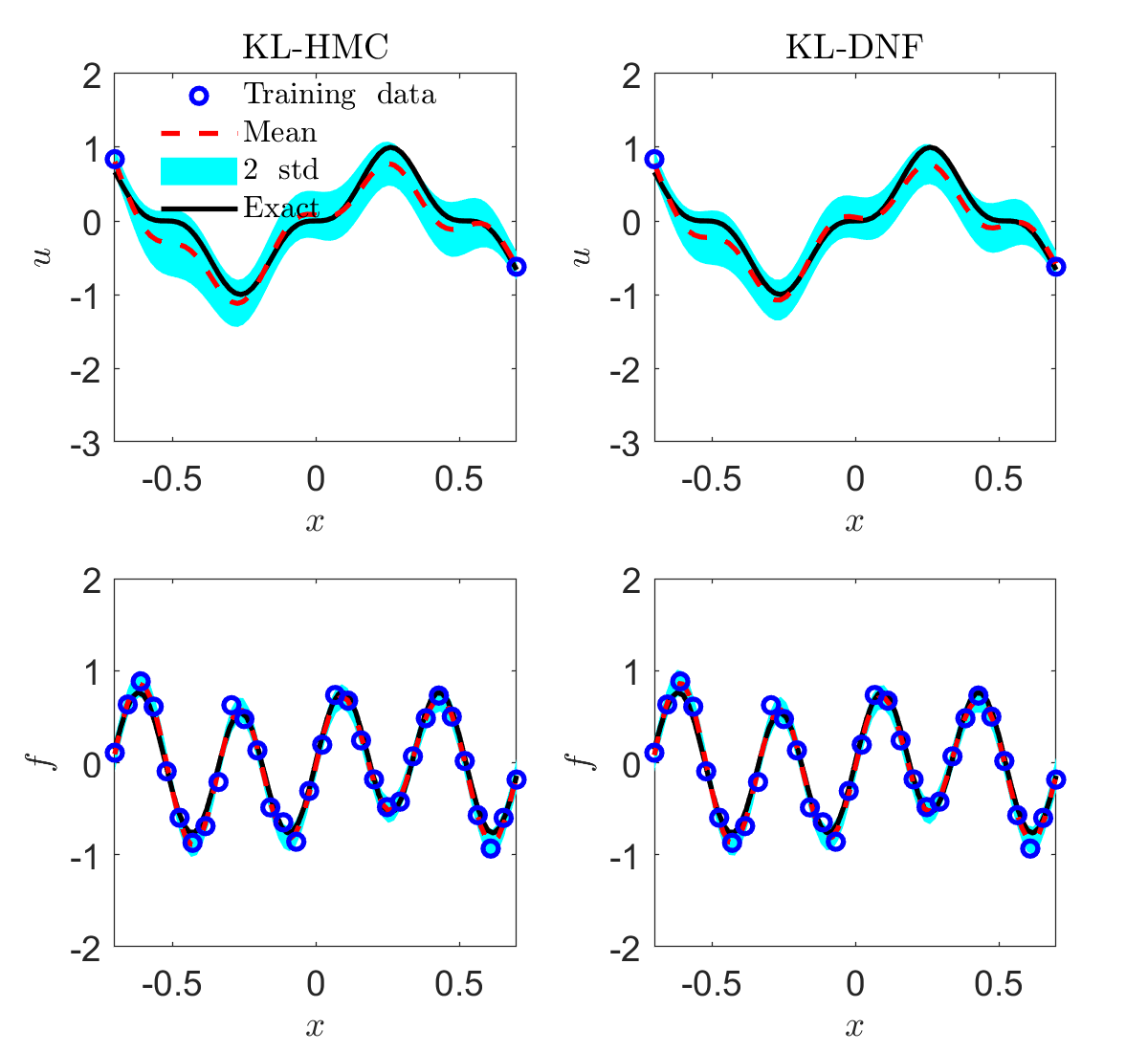}}
    \caption{1D nonlinear Poisson equation (KL) - forward problem: Predicted $u$ and $f$ with two data noise scales.   (a): $\epsilon_f \thicksim \mathcal{N}(0, 0.01^2)$, $\epsilon_b \thicksim \mathcal{N}(0, 0.01^2)$. (b): $\epsilon_f \thicksim \mathcal{N}(0, 0.1^2)$, $\epsilon_b \thicksim \mathcal{N}(0, 0.1^2)$.
    }
    \label{fig:klforward}
\end{figure}

\begin{figure}
    \centering
    \subfigure[]{
    \includegraphics[width=0.45\textwidth]{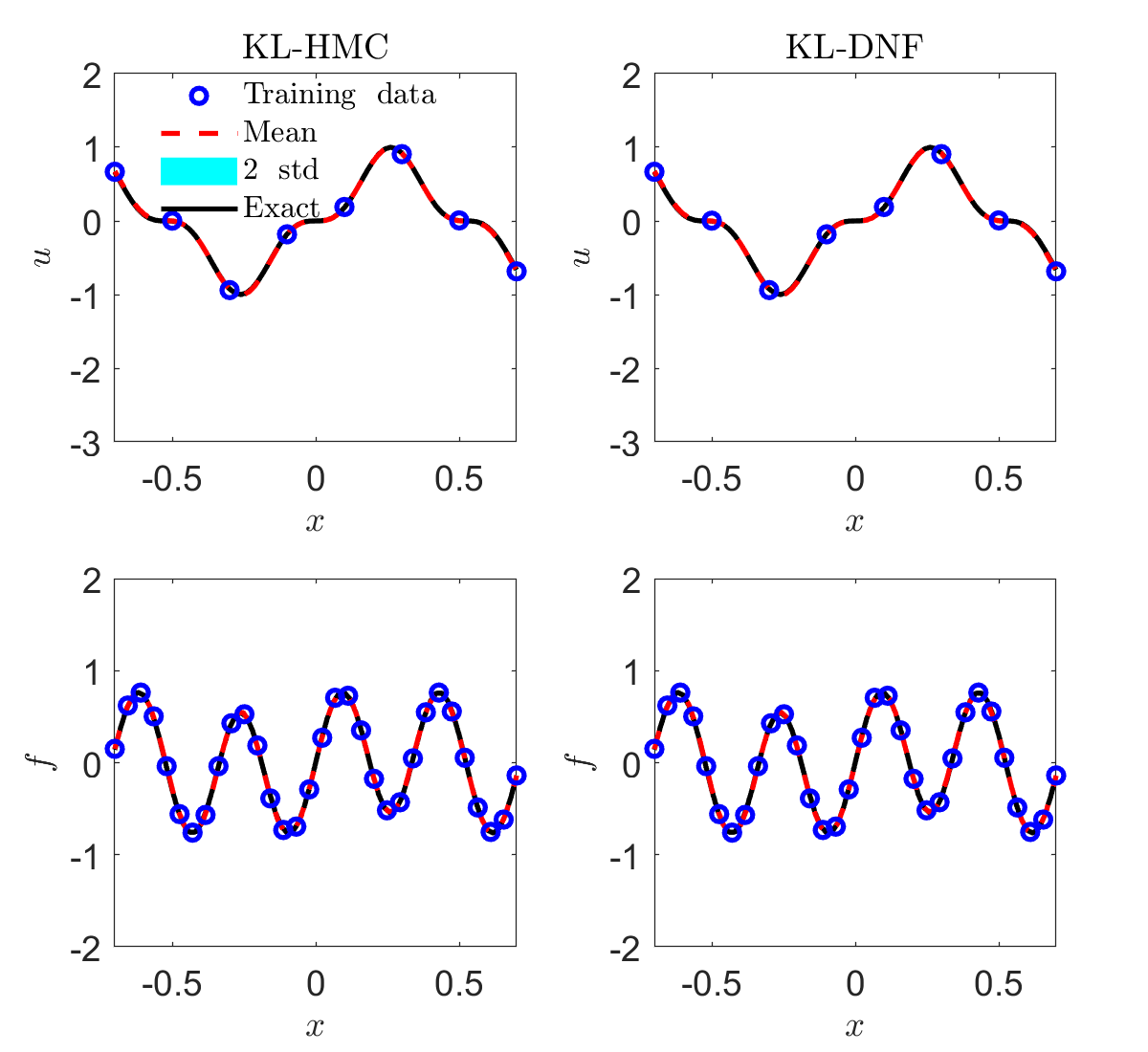}}
    \subfigure[]{
    \includegraphics[width=0.45\textwidth]{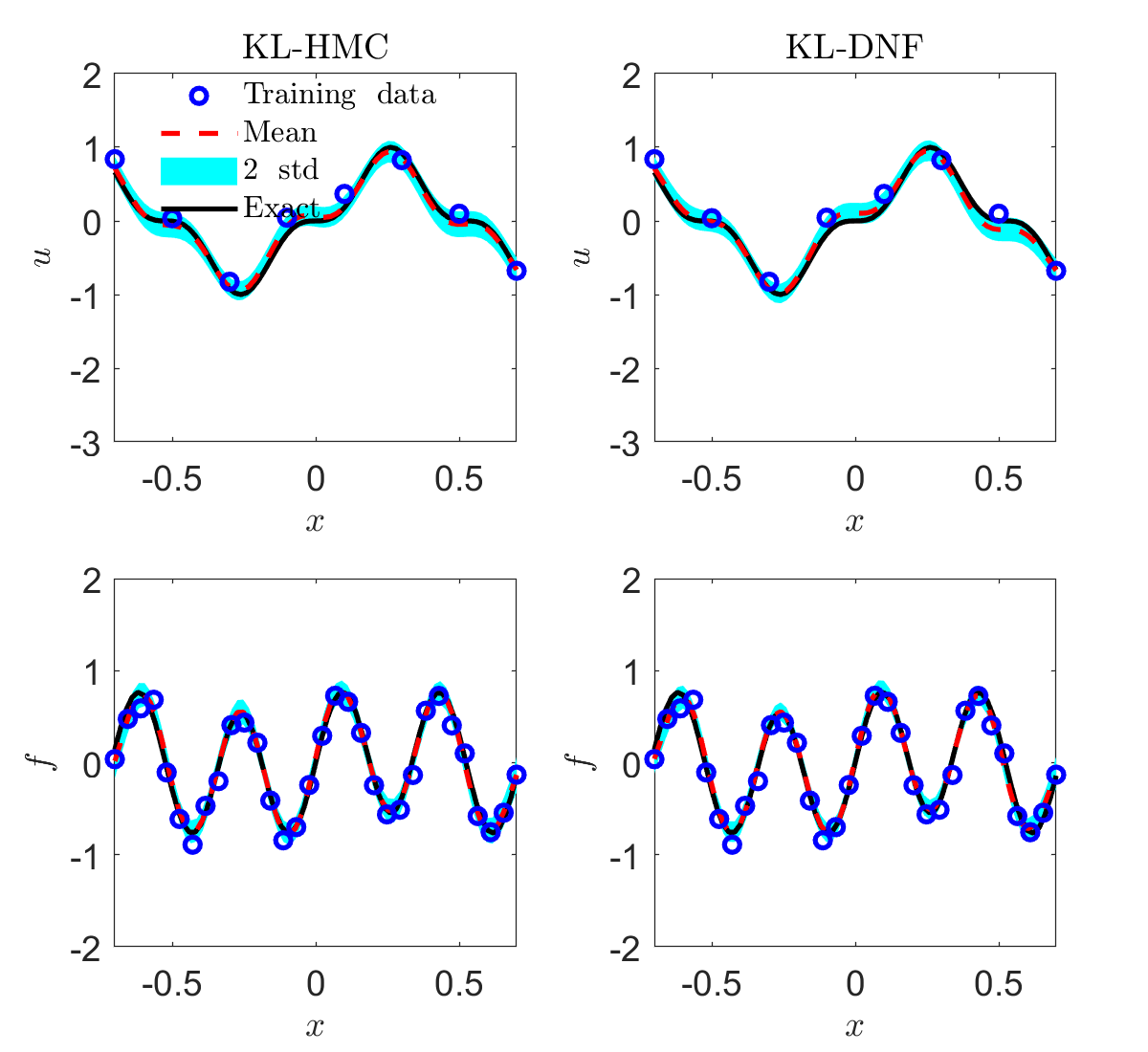}}
    \caption{1D diffusion-reaction system with nonlinear source term (KL): Predicted $u$ and $f$ with two data noise scales.   (a) $\epsilon_f \thicksim \mathcal{N}(0, 0.01^2)$, $\epsilon_u \thicksim \mathcal{N}(0, 0.01^2)$, $\epsilon_b \thicksim \mathcal{N}(0, 0.01^2)$. (b) $\epsilon_f \thicksim \mathcal{N}(0, 0.1^2)$, $\epsilon_u \thicksim \mathcal{N}(0, 0.1^2)$, $\epsilon_b \thicksim \mathcal{N}(0, 0.01^2)$.
    }
    \label{fig:klinverse}
\end{figure}

The predicted $u$ and $f$ are illustrated in Figs.~\ref{fig:klforward}-\ref{fig:klinverse}. The results from the KL-HMC and KL-DNF are almost the same, and the predicted means for $u$ and $f$ are close to the exact solutions. In addition, the predicted $k$ is displayed in Table \ref{table:KLkpred}. Similarly, the predicted means for both cases fit the exact solution quite well. Furthermore, we also note that (1) the standard deviation increases with the increasing noise scale, and (2) the errors are bounded by two standard deviations. All the results are similar as those from the B-PINNs presented in Sec. \ref{sec:1Dnonforward} and Sec.~\ref{sec:inpde}.

\begin{table}
\centering
{\footnotesize
\begin{tabular}{cc|cc}
\hline \hline
  \multicolumn{2}{c|}{ Noise scale}  & {KL-HMC} & { KL-DNF}  \\
  \hline
  \multirow{2}{*}{0.01} &Mean  & 0.706 & 0.705 \\
  & Std   &  {\footnotesize $5.63 \times 10^{-3}$} & {\footnotesize $2.63 \times 10^{-3}$} \\
  \multirow{2}{*}{0.1} &Mean   & 0.694 & 0.709  \\
  & Std  & {\footnotesize $5.82 \times 10^{-2}$} & {\footnotesize $5.21 \times 10^{-2}$} \\
  \hline \hline
\end{tabular}
}
\caption{1D diffusion-reaction system with nonlinear source term (KL): Predicted mean and standard deviation for $k$ using KL. The exact solution for $k$ is $0.7$. }
\label{table:KLkpred}
\end{table}

As for the computational cost, the DNF takes about one day to finish the training for the 1D diffusion-reaction problem, while it takes about 4 mins for the HMC. Although the DNF is computationally much more expensive than the HMC, we would like to remark that upon completion of the training, it is more convenient to draw \textit{independent} samples from the target distribution using DNF compared with HMC. This strength of DNF has no significant benefit in current work, but could be helpful for other tasks.

Here, we also conduct a brief comparison on the computational cost between the KL-HMC and the B-PINN-HMC based on the 1D diffusion-reaction problem. Due to the relatively small number of parameters in the truncated KL expansion, in the 1D test cases, KL-HMC takes much less time than B-PINN-HMC. In particular, KL-HMC takes about 4 mins compared to about 20 mins for B-PINN-HMC. However, we remark that the truncated KL expansion would suffer from the ``curse of dimensionality'' when approximating high dimensional functions, while deep neural networks are known to be efficient for high-dimensional function approximation \cite{cheridito2019efficient}.

\section{Summary}
\label{sec:summary}

There are many sources of uncertainty in data-driven PDE solvers, including \textit{aleatoric uncertainty} associated with noisy data, \textit{epistemic uncertainty} associated with unknown parameters, and \textit{model uncertainty} associated with the type of PDE that models the target phenomena. In this paper, we address aleatoric uncertainty for solving forward and inverse PDE problems, based on noisy data associated with the solution, source terms and boundary conditions. In particular, we employ physics-informed neural networks (PINNs) to solve PDEs, using automatic differentiation, with the accuracy of the solution depending critically on the quality of the training data.

In order to quantify uncertainty and improve the accuracy of PINNs, we propose a general Bayesian framework, consisting of a Bayesian neural network for the solution, subject to the PDE constraint that serves as a prior, combined with different estimators for the posterior, namely, the Hamiltonian Monte Carlo (HMC) method and the variational inference (VI). We conduct a comprehensive comparison among different methods, i.e., the B-PINN with HMC, B-PINN with VI, and PINN with dropout, which is also used to quantify the uncertainty of neural networks.  We investigate both linear and nonlinear PDEs with noisy data. Our experiments demonstrate  good accuracy and robustness of B-PINN-HMC, but B-PINN-VI usually gives unreasonable uncertainties, which could be attributed to the fact  that the posterior distribution is approximated by a factorizable Gaussian distribution. Moreover, dropout which is not based on the Bayesian framework can hardly provide satisfactory uncertainty quantification, in agreement with  \cite{yao2019quality}. In addition,  we also compare the performance of the B-PINN-HMC with the PINNs. The results show that PINNs could easily overfit the noisy data and get less accurate results than B-PINN-HMC. 

As an alternative surrogate model, we replace the BNN with a truncated KL expansion and combine it with HMC or deep normalizing flow (DNF) models for estimating the posterior. We repeated some of the experiments and found that both KL-HMC and KL-DNF yield equally accurate results as B-PINN-HMC, but at a reduced cost for KL-HMC. This KL-based Bayesian framework could also be very effective in uncertainty quantification of data-driven PDE solvers, but is limited to low dimensional problems. We explored the possibility of DNF as a posterior estimator in the KL-based Bayesian framework. While much more computationally expensive than HMC, upon completion of training, DNF can draw \textit{independent} samples more easily from the target distribution. This strength of DNF has no significant benefit in the current Bayesian framework, but could be helpful for other tasks.

While the choice of priors for B-PINNs may have a significant influence on the posterior predictions especially in the cases with small data, such choice of priors, including the structure of neural networks and the prior distribution for the parameters, remains an open problem. Also, in the current work, we only tested the cases where the data size is up to several hundreds; for the big data case, we may need to use other posterior sampling methods in conjunction with mini-batch techniques, like stochastic HMC \cite{chen2014stochastic,ding2014bayesian,ma2015complete}, which needs further investigation in the future.

\section*{Acknowledgement}
This work was supported by the PhILMS grant DE-SC0019453, the DARPA-AIRA grant HR00111990025 and the NIH-Yale grant U01 HL142518.

\appendix
\section{BNNs with different priors}
\label{sec:prior}

\begin{figure}[H]
    \centering
    \subfigure[]{\label{fig:priora}
    \includegraphics[width=0.3\textwidth]{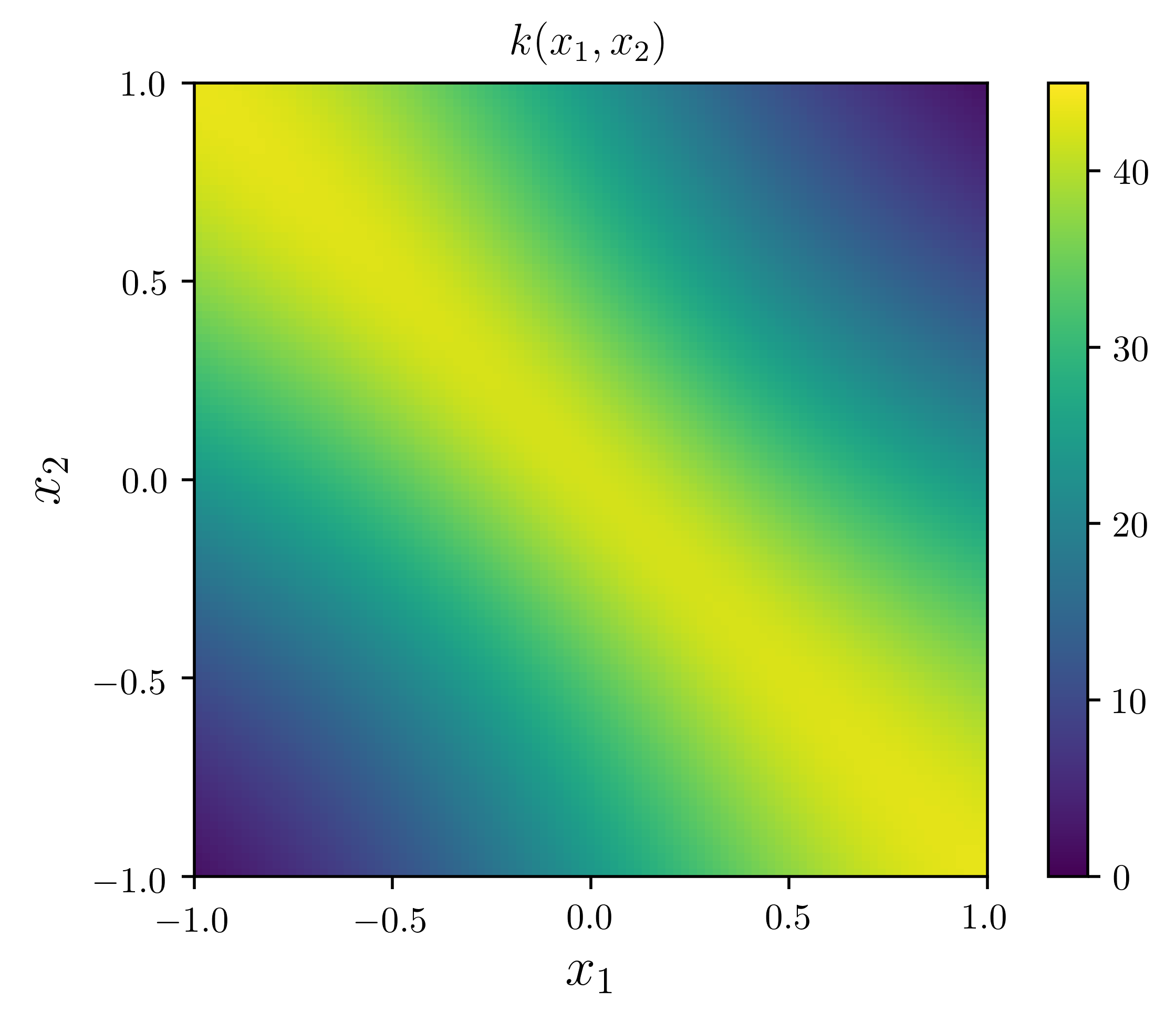}}
    \subfigure[]{\label{fig:priorb}
    \includegraphics[width=0.3\textwidth]{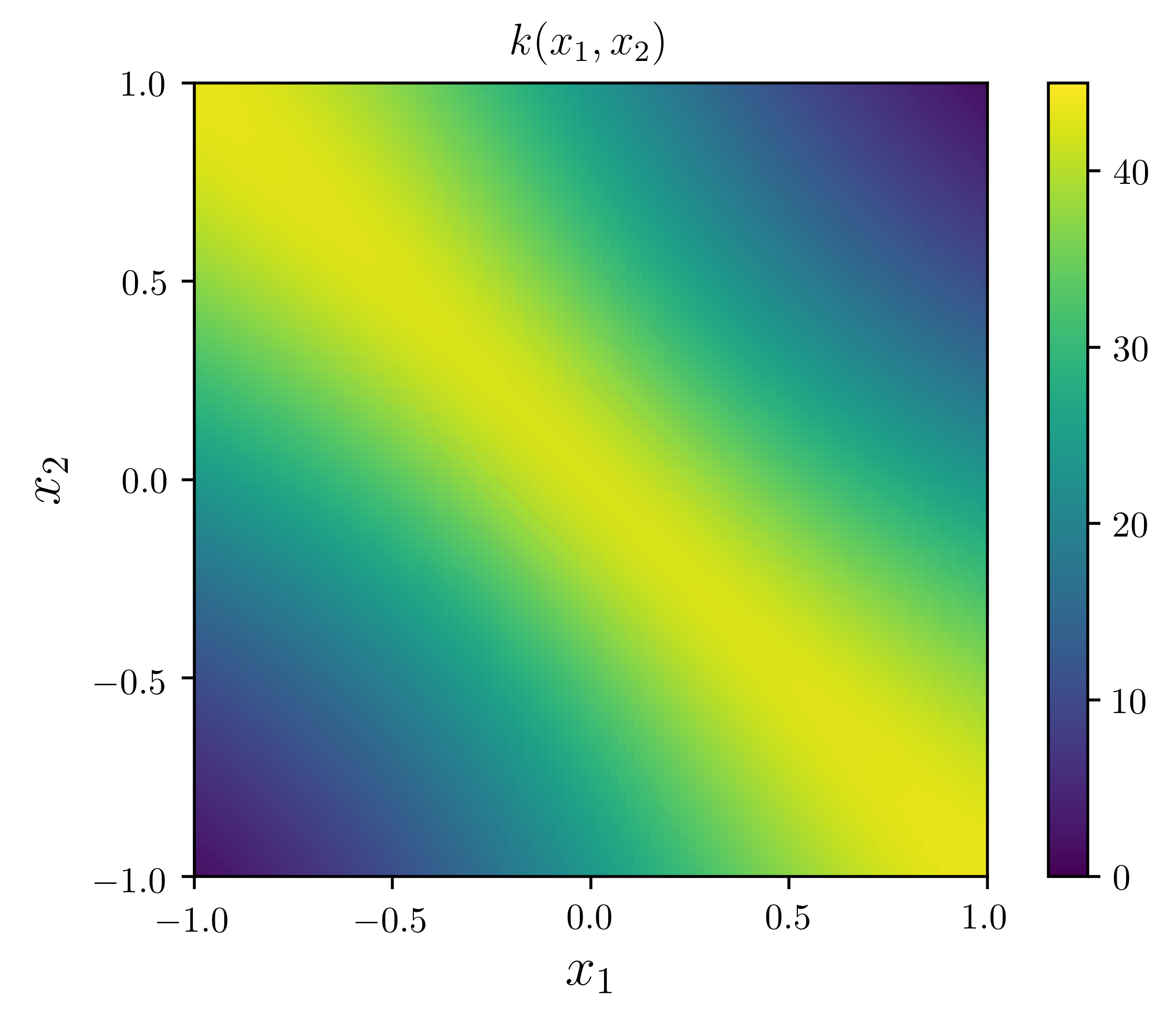}}
    \subfigure[]{\label{fig:priorc}
    \includegraphics[width=0.3\textwidth]{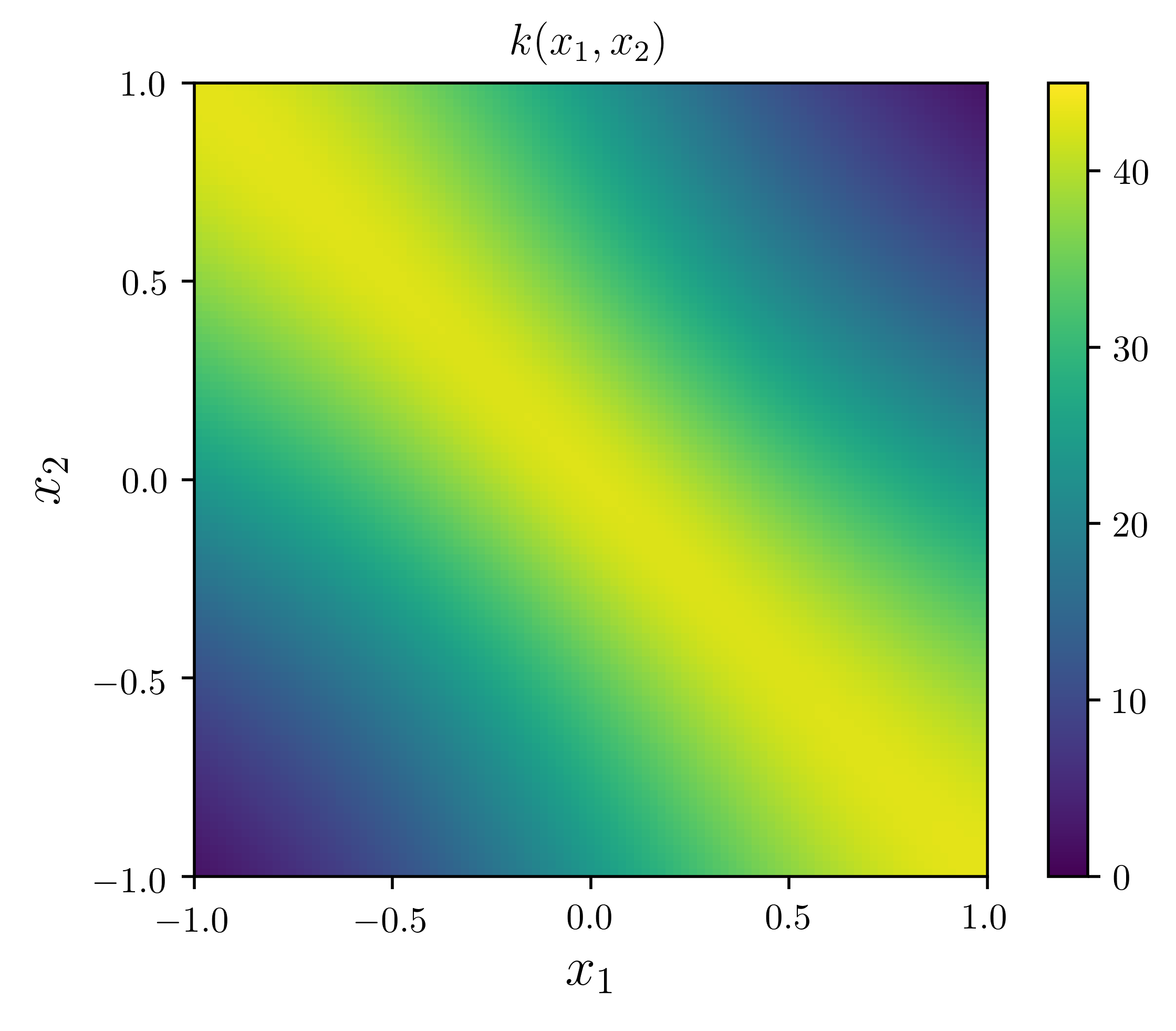}} \\
    \subfigure[]{\label{fig:priord}
    \includegraphics[width=0.3\textwidth]{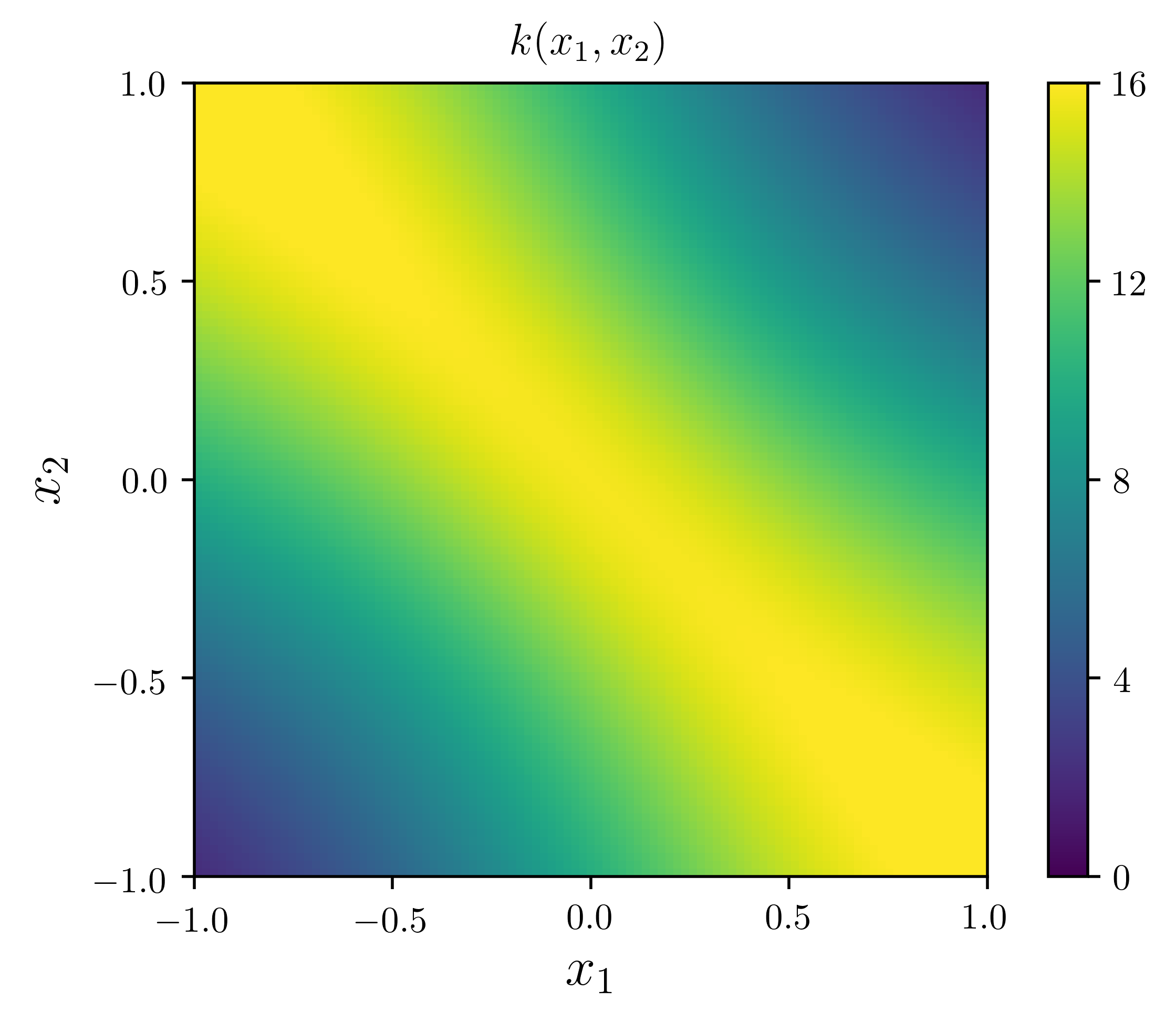}}
    \subfigure[]{\label{fig:priore}
    \includegraphics[width=0.3\textwidth]{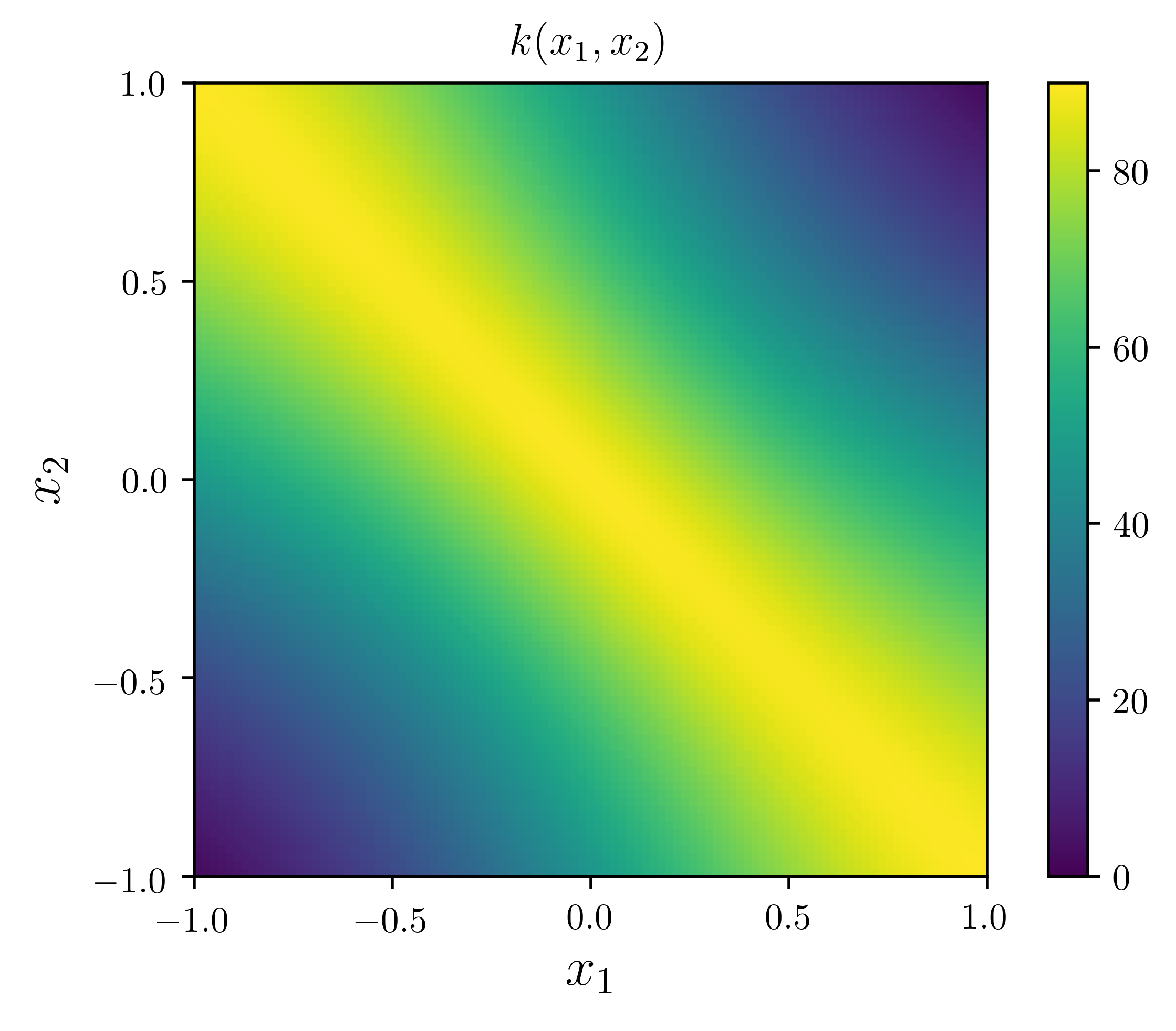}}
    \caption{Covariance functions $k(x_1,x_2) = cov(\tilde{u}(x_1),\tilde{u}(x_2))$ for BNNs with different architectures. $L = 2, \sigma_{b,l}= 1$ for $l=0,1,2$ in all the cases. (a) $N_1 = N_2 = 20$, $\sigma_{w,0} = 1.0$, $\sigma_{w,1}= \sigma_{w,2}= \sqrt{5/2}$ (b) $N_1 = N_2 = 50$,  $\sigma_{w,0}=\sigma_{w,1}= \sigma_{w,2}= 1.0$, (c) $N_1 = N_2 = 100$,  $\sigma_{w,0}=1.0$, $\sigma_{w,1}= \sigma_{w,2}= \sqrt{1/2}$. (d) $N_1 = N_2 = 20$, $\sigma_{w,0}=\sigma_{w,1}= \sigma_{w,2}= 1.0$, (e) $N_1 = N_2 = 100$, $\sigma_{w,0}=\sigma_{w,1}= \sigma_{w,2}= 1.0$.}
    \label{fig:prior}
\end{figure}

The posterior distribution depends on both the prior and the observed data in the Bayesian framework. Given the same observation, surrogate models with similar prior distributions should also provide similar posterior distributions.  For the cases of input dimension $N_x =1$, we illustrate the covariance functions for five representative priors in Fig.~\ref{fig:prior}, which are estimated from $100,000$ independent samples of neural network parameters drawn from the prior. Note that $\sqrt{N_l}\sigma_{w,l}$ is fixed for cases (a), (b) and (c), and we can see that the covariance functions are similar for the three cases.

\begin{figure}
    \centering
    \subfigure[]{\label{fig:bnns_priora}
    \includegraphics[width = 0.23\textwidth]{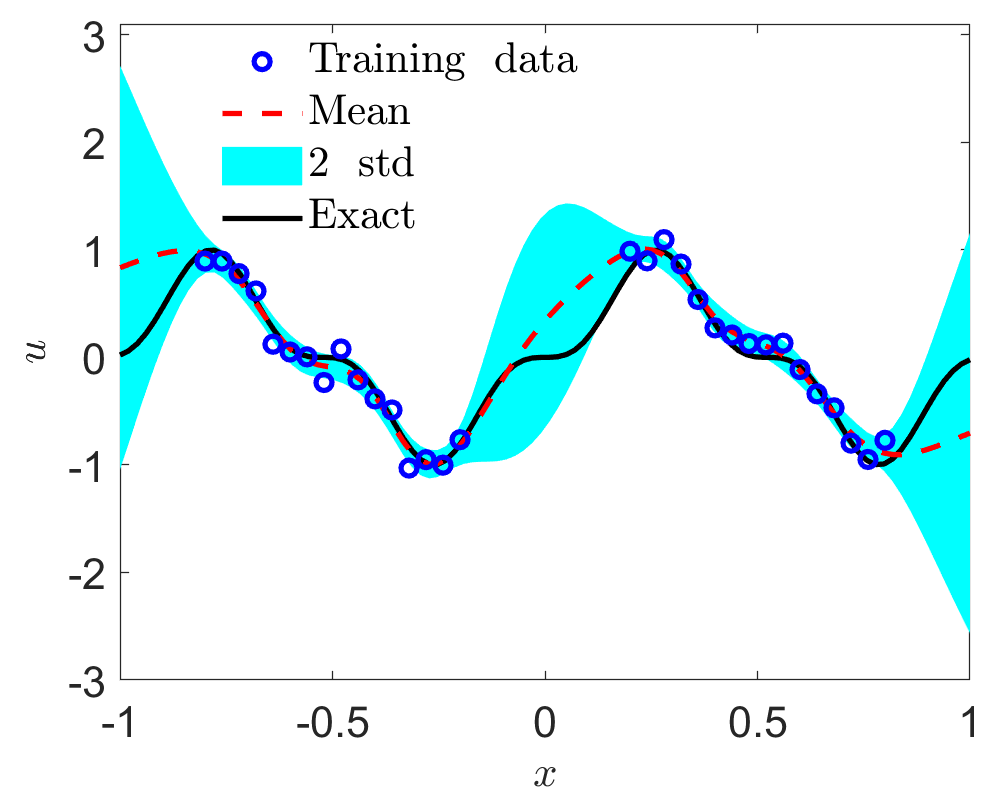}}
    \subfigure[]{\label{fig:bnns_priorb}
    \includegraphics[width = 0.23\textwidth]{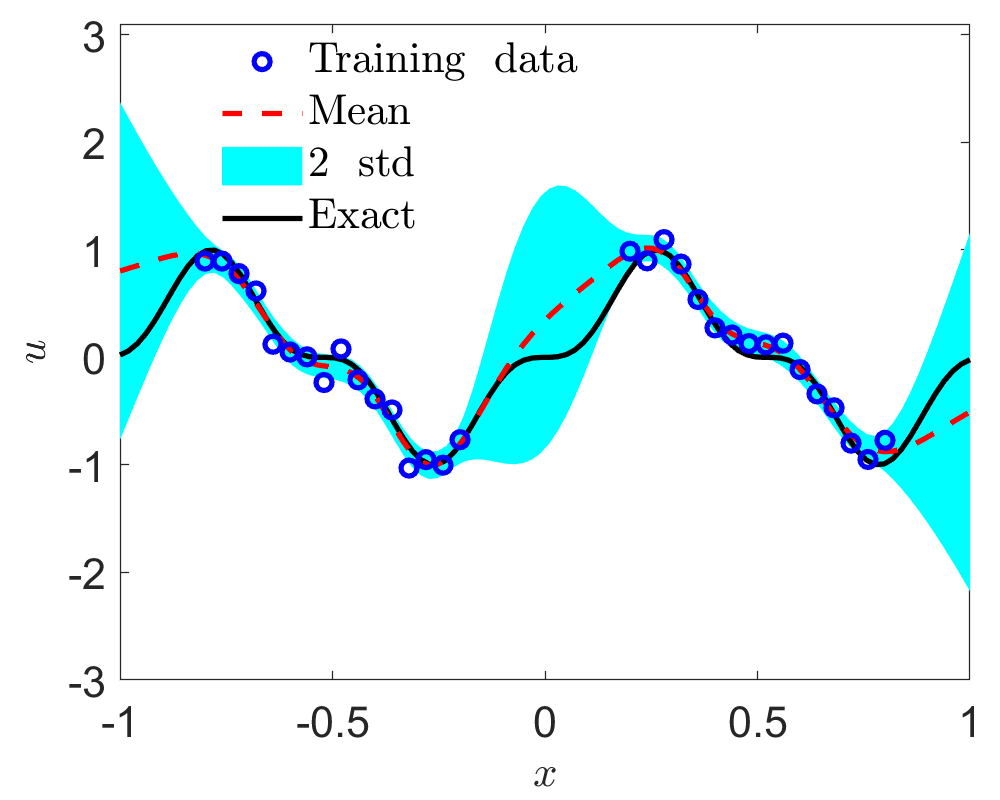}}
    \subfigure[]{\label{fig:bnns_priorc}
    \includegraphics[width = 0.23\textwidth]{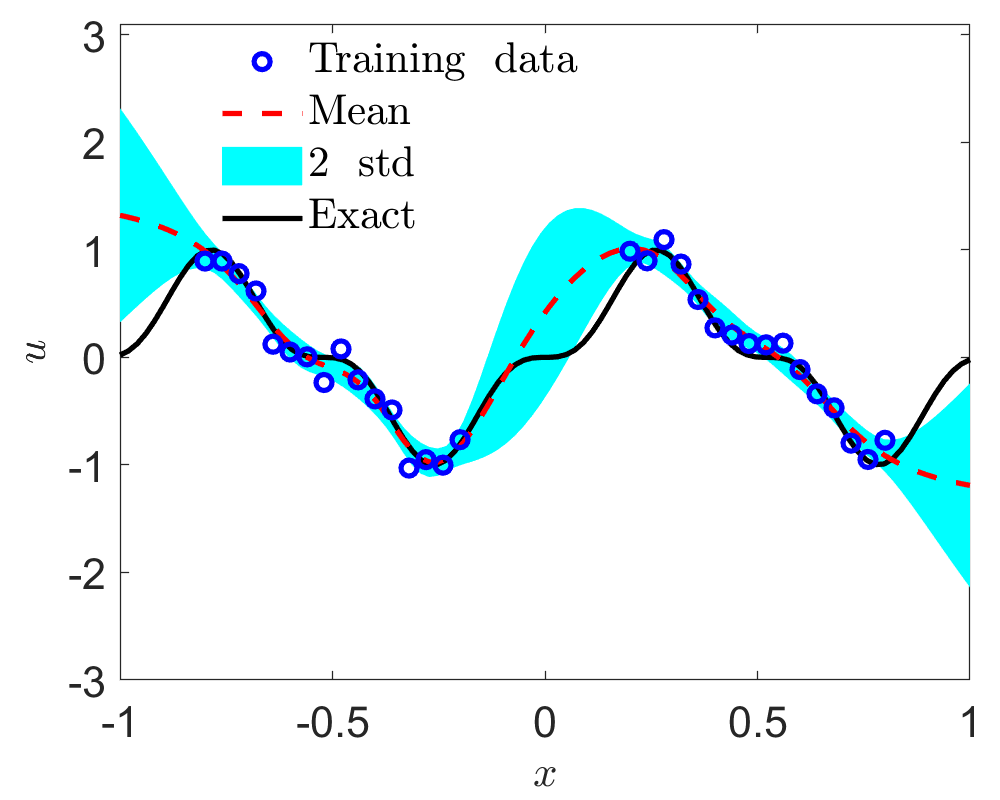}}
    \subfigure[]{\label{fig:bnns_priord}
    \includegraphics[width = 0.23\textwidth]{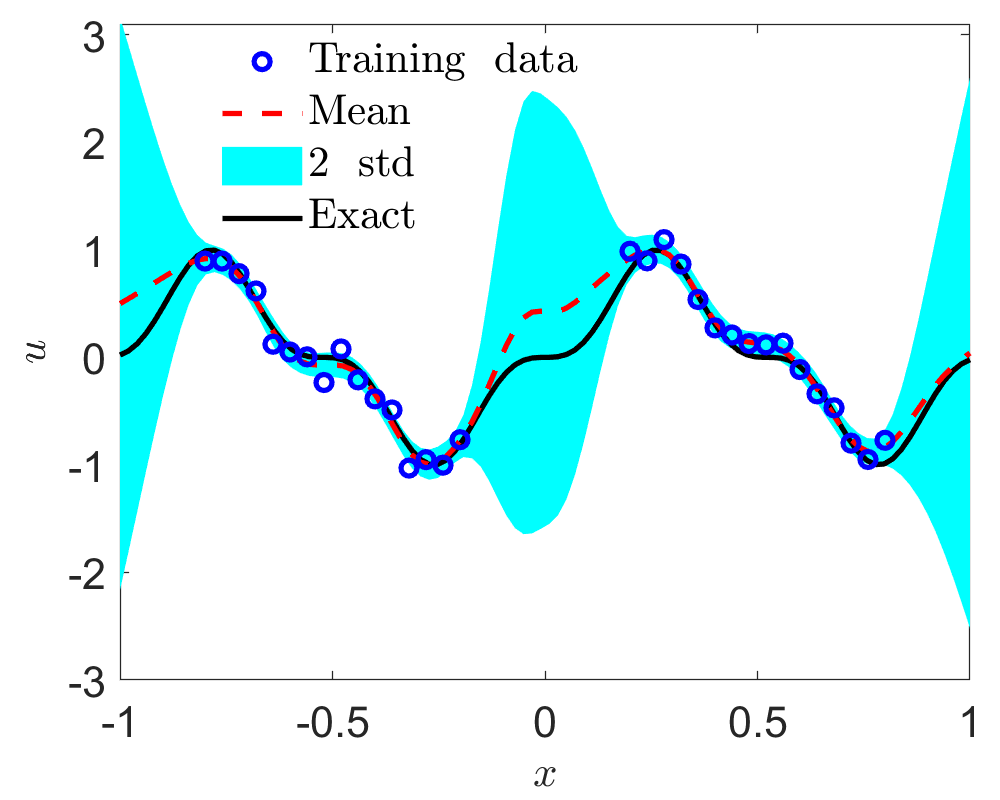}}
    \caption{BNN-HMC with different priors for function approximation. 
    (a) $L = 2, N_1 = N_2 = 20$, $\omega_1 = \omega_2 \thicksim \mathcal{N} (0, \sqrt{5/2})$,  
    (b) $L = 2, N_1 = N_2 = 100$, $\omega_1 = \omega_2 \thicksim \mathcal{N} (0, \sqrt{1/2})$,
    (c) $L = 2, N_1 = N_2 = 20$, $\omega_1 = \omega_2 \thicksim \mathcal{N} (0, 1)$, 
    (d) $L = 2, N_1 = N_2 = 100$, $\omega_1 = \omega_2 \thicksim \mathcal{N} (0, 1)$. }
    \label{fig:bnns_prior}
\end{figure}

We plot the results for approximating the same function in Eq. \eqref{eq:func} with same data, using BNNs with different architectures in Fig. \ref{fig:bnns_prior}. The predicted means and standard deviations are observed to be quite similar for cases in Figs. A.\ref{fig:bnns_priora}-A.\ref{fig:bnns_priorc}, which is consistent with the fact that the covariance functions of the priors for these three cases are similar (Figs. A.\ref{fig:priora}-A.\ref{fig:priorc}). In addition,  the results in Figs. A.\ref{fig:bnns_priorc}-A.\ref{fig:bnns_priord} are different from those in Fig. \ref{fig:funce}, which is not surprising since their covariance functions are totally different (Fig. \ref{fig:prior}).

\section{Deep Normalizing Flow Models}
\label{sec:DNF}

Deep normalizing flow (DNF) models provide a powerful mechanism for sampling from a wide range of probability distributions.  While there have been many versions of normalizing flow models, as a demonstrating example, in this paper we use a potential flow to build the bijective transformation. We refer the readers to \cite{yang2019potential,zhang2018monge} for similar approaches. In particular, the bijective transformation $G$ is defined as the map from $\boldsymbol{u}$ at time $t=0$ to $T$ of the following ODE:
\begin{equation}
\label{eqn:mapODE}
    \begin{aligned}
        \frac{d\boldsymbol{u}}{dt} = \nabla \varphi(\boldsymbol{u}, t; \boldsymbol{\zeta}),
    \end{aligned}
\end{equation}
where $\varphi(\boldsymbol{u}, t; \boldsymbol{\zeta})$ is represented by a deep neural network with parameter $\boldsymbol{\zeta}$, which takes the concatenation of $\boldsymbol{u}$ and $t$ as input, and outputs a real number.
Consequently, we have the following ODE for the probability density:
\begin{equation}
\label{eqn:densityODE}
    \begin{aligned}
        \frac{d\ln P(\boldsymbol{u}(t),t)}{dt} = - \nabla^2 \varphi(\boldsymbol{u}, t; \boldsymbol{\zeta}),
    \end{aligned}
\end{equation}
 where $P(\boldsymbol{u},0) = P_{\mu_I}(\boldsymbol{u})$ is the density of $\mu_I$ at $\boldsymbol{u}$, and  $P(\boldsymbol{u},T) = P_{\mu_O}(\boldsymbol{u})$ is the density of $\mu_O$ at $\boldsymbol{u}$. 
 
 Here, we use the forward Euler scheme to solve the ODE (\ref{eqn:mapODE}). Suppose the time step is $\delta t = T/n$, then
 \begin{equation}
     \begin{aligned}
     \boldsymbol{u}_0(\boldsymbol{z}) &= \boldsymbol{z}, \\
          \boldsymbol{u}_i(\boldsymbol{z}) = \boldsymbol{u}_{i-1}(\boldsymbol{z})  &+ \delta t \nabla \varphi(\boldsymbol{u}_{i-1}(\boldsymbol{z}), \frac{(i-1)T}{n}; \boldsymbol{\zeta}), \quad i = 1,2...n \\
     \end{aligned}
 \end{equation}
so that $G(\boldsymbol{z}) = \boldsymbol{u}_{n}(\boldsymbol{z})$. Similarly, we have the forward Eular scheme for ODE (\ref{eqn:densityODE}):
\begin{equation}
\begin{aligned}
\ln P_0(\boldsymbol{u}_0(\boldsymbol{z})) &= \ln P_{\mu_I}(\boldsymbol{z}), \\
         \ln P_i(\boldsymbol{u}_i(\boldsymbol{z})) =  \ln P_{i-1}(\boldsymbol{u}_{i-1}(\boldsymbol{z}))  &- \delta t \nabla^2 \varphi(\boldsymbol{u}_{i-1}(\boldsymbol{z}), \frac{(i-1)T}{n}; \boldsymbol{\zeta}), \quad i = 1,2...n \\
\end{aligned}
\end{equation}
so that $\ln P(G(\boldsymbol{z}), T) = \ln P_{n}(\boldsymbol{u}_{n}(\boldsymbol{z}))$.

For our problems, where the target distribution $\nu$ is given by the posterior density $P(\boldsymbol{\theta}|\mathcal{D})$, we tune the parameters in $\varphi$ to minimize 
\begin{equation}
    \begin{aligned}
        F(\mu_O, \nu) &= D_{KL}(\mu_O||\nu) \\
                      &= \mathbb{E}_{\boldsymbol{\theta}\sim \mu_O}[\ln P(\boldsymbol{\theta}, T) - \ln P(\boldsymbol{\theta} | \mathcal{D})] \\
                      &\simeq \mathbb{E}_{\boldsymbol{\theta}\sim \mu_O}[\ln P(\boldsymbol{\theta}, T) - \ln P(\boldsymbol{\theta}) - \ln P(\mathcal{D}|\boldsymbol{\theta})] \\
                      &= \mathbb{E}_{\boldsymbol{z}\sim \mu_I}[\ln P(G(\boldsymbol{z}), T) - \ln P(G(\boldsymbol{z})) - \ln P(\mathcal{D}|G(\boldsymbol{z}))], \\
    \end{aligned}
\end{equation}
where $D_{KL}$ represents the Kullback-Leibler  divergence, and ``$\simeq$'' represents equality up to a constant.  In this paper we employ the Adam optimizer to train $\bm{\zeta}$.

Ideally, $\mu_O$ and $\nu$ would be sufficiently close to each other after the convergence of $F(\mu_O, \nu)$. We could then sample $\{\boldsymbol{z}^{(j)}\}_{j=1}^M$ from $\mu_{I}$, and get statistics of $P(\bm{\theta}|\mathcal{D})$ from $\{G(\boldsymbol{z}^{(j)})\}_{j=1}^M$. 

The detailed algorithm is given in Algorithm \ref{alg:flow}.

\begin{algorithm}[H]
\caption{Normalizing Flow}
\label{alg:flow}
\begin{algorithmic}
\Require an initial state for $\bm{\zeta}$.
\For{$k=1,2...N$}
\State Sample $\{\boldsymbol{z}^{(j)}\}_{j=1}^{N_z}$ independently from $\mu_I$. \;
\State $L(\boldsymbol{\zeta}) \leftarrow \frac{1}{N_z}\sum_{j=1}^{N_z} [\ln P(G(\boldsymbol{z}^{(j)}), T) - \ln P(G(\boldsymbol{z}^{(j)})) - \ln P(\mathcal{D}|G(\boldsymbol{z}^{(j)}))]$. \;
\State Update $\boldsymbol{\zeta}$ with gradient $\nabla_{\boldsymbol{\zeta}}L(\boldsymbol{\zeta})$ using Adam optimizer. \;
\EndFor
\State Sample $\{\boldsymbol{z}^{(j)}\}_{j=1}^{M}$ independently from $\mu_{I}$. \;
\State Calculate $\{ \tilde{u}(\boldsymbol{x},G(\boldsymbol{z}^{(j)}))\}_{j=1}^M$ as samples of $u(\boldsymbol{x})$, similarly for other terms. \;
\end{algorithmic}
\end{algorithm}

In this paper, we set time span $T = 1$, and time steps in the forward Euler scheme $n = 50$ in the forward problems, while $n = 10$ in the inverse problems. The neural networks for $\varphi$ have 3 hidden layers, each of width 128. For all the cases, the total training steps $N = 100,000$ and batch size $N_z = 16$. The hyperparameters for the Adam optimizer are set as $l = 10^{-4}, \beta_1 = 0.9, \beta_2 = 0.999$.

\bibliographystyle{elsarticle-num-names}
\bibliography{B-PINNs.bib}

\end{document}